\documentclass[sigconf]{acmart}
\usepackage{algorithm}
\usepackage{algpseudocode}
\usepackage{multirow}
\usepackage{bm}
\usepackage{pifont}
\usepackage{amsmath}
\usepackage{enumitem}
\usepackage{xspace}
\newcommand{\method}{\textsc{CoBAD}\xspace}

\newcommand{\R}{\mathbb{R}}

\newcommand{\Z}{{\bm Z}}
\newcommand{\z}{{\bm z}}
\newcommand{\F}{\mathcal{F}}

\newcommand{\G}{\mathcal{G}}
\newcommand{\V}{\mathcal{V}}
\newcommand{\Edge}{\mathcal{E}}

\newcommand{\NUMOSIM}{{\small{\textsf{NUMOSIM}}}\xspace}

\newcommand{\trialfourone}{{\small{\textsf{MobilitySim-A}}}\xspace}
\newcommand{\trialfourtwo}{{\small{\textsf{MobilitySim-B}}}\xspace}

\usepackage{xcolor}

\newcommand{\blue}[1]{\textcolor{blue}{#1}}
\usepackage{hyperref}
\newcommand{\figref}[1]{\textcolor{blue}{Figure~\ref{#1}}}
\newcommand{\tabref}[1]{\textcolor{purple}{Table~\ref{#1}}}
\newcommand{\eqnref}[1]{\textcolor{teal}{Eq.~(\ref{#1})}}
\newcommand{\appref}[1]{\textcolor{olive}{Appx. ~\ref{#1}}}

\AtBeginDocument{%
  }

\setcopyright{acmlicensed}
\copyrightyear{2018}
\acmYear{2018}
\acmDOI{XXXXXXX.XXXXXXX}
\acmConference[ACM SIGSPATIAL'25]{Make sure to enter the correct
  conference title from your rights confirmation email}{November 3--6,
  2025}{Minneapolis, MN, USA}
\acmISBN{978-1-4503-XXXX-X/2018/06}

\begin{document}

% \title{Towards Collective Anomaly Detection for Human Mobility}

\title[Modeling Collective Behaviors for Human Mobility Anomaly Detection]{\textsc{CoBAD}: Modeling Collective Behaviors for Human Mobility \\ Anomaly Detection}


\author{Haomin Wen}
% \authornote{Equal contribution}
\affiliation{%
  \institution{Carnegie Mellon University}
 % \city{Pittsburgh}
   \country{Pittsburgh,USA}
  }
\email{haominwe@andrew.cmu.edu}
\author{Shurui Cao}
\affiliation{%
  \institution{Carnegie Mellon University}
 % \city{Pittsburgh}
  \country{Pittsburgh, USA}
 }
\email{shuruic@andrew.cmu.edu}

\author{Leman Akoglu}
\affiliation{%
  \institution{Carnegie Mellon University}
  %\city{Pittsburgh}
  \country{Pittsburgh, USA}
  }
\email{lakoglu@andrew.cmu.edu}

\renewcommand{\shortauthors}{Haomin Wen et al.}

\begin{abstract}
  Detecting anomalies in human mobility is essential for applications such as public safety and urban planning. While traditional anomaly detection methods primarily focus on individual movement patterns (e.g., a child should stay at home at night), collective anomaly detection aims to identify irregularities in collective mobility behaviors across individuals (e.g., a child is at home alone while the parents are elsewhere) and remains an underexplored challenge. Unlike individual anomalies, collective anomalies require modeling spatiotemporal dependencies between individuals, introducing additional complexity. To address this gap, we propose \method, 
  %\lmn{shall we change the name to CoBAD? B for behavior. Anomaly~BAD?} 
%  the first-ever \lmn{we shouldnt claim to be first} 
 a novel model designed to capture \underline{Co}llective \underline{B}ehaviors for human mobility \underline{A}nomaly \underline{D}etection. 
  We first formulate the problem as unsupervised learning over Collective Event Sequences (CES) with a co-occurrence event graph, where CES represents the event sequences of related individuals. \method then employs a two-stage attention mechanism to model both the individual mobility patterns and the interactions across multiple individuals. 
  Pre-trained on large-scale collective behavior data through masked event and link reconstruction tasks, \method is able to detect two types of collective anomalies: unexpected co-occurrence anomalies and absence anomalies, the latter of which has been largely overlooked in prior work.
  % by massive collective behaviors with the goal of reconstructing randomly masked events and their associated links, \method poses the ability to detect two types of collective anomalies: unexpected co-occurrence anomaly, as well as absence anomaly (being largely ignored in existing approaches).
  Extensive experiments on large-scale mobility datasets demonstrate that \method significantly outperforms existing anomaly detection baselines, achieving an improvement of 13\%-18\% in AUCROC and 19\%-70\% in AUCPR. %\lmn{AUCPR so it is consistent with AUCROC}. 
  All source code is available at 
\href{https://github.com/wenhaomin/CoBAD}{\color{purple}{https://github.com/wenhaomin/CoBAD}}.

 %  Based on this, we elaborately design a collective anomaly detection score via reconstructed node features and links at the inference stage. 

  %   At last, we train the model via masked pretraining, which randomly masks events and links associated with the masked events.

  %  with asynchronous temporal occurrences
  
\end{abstract}

% We first formulate the problem as learning on Collective Event Sequences (CES) with an event-graph, where CES represents mobility event sequences with asynchronous temporal occurrences, and the event-graph models co-location and co-occurrence dependencies among individuals. \method then leverages a Two-Stage Attention (TSA) mechanism, consisting of cross-time attention to model sequential mobility behaviors and cross-people attention to encode dynamic interactions between individuals. We employ an unsupervised learning framework based on masked event prediction and link reconstruction, based on which we propose a collective anomaly detection score via reconstructed node and edge features at the inference stage. Extensive experiments on large-scale mobility datasets demonstrate that \method significantly outperforms existing anomaly detection baselines, achieving state-of-the-art performance in identifying both individual and collective anomalies. 

% Temporal-order Asynchronous Sequences

\begin{CCSXML}
<ccs2012>
   <concept>
       <concept_id>10002951.10003227.10003236</concept_id>
       <concept_desc>Information systems~Spatial-temporal systems</concept_desc>
       <concept_significance>500</concept_significance>
       </concept>
 </ccs2012>
\end{CCSXML}

\ccsdesc[500]{Information systems~Spatial-temporal systems}


% \keywords{Human Mobility, Collective Behaviors, Anomaly Detection}




\maketitle

\section{Introduction}

\par
\begin{figure}[!t]
    \centering
    \includegraphics[width=1 \linewidth]{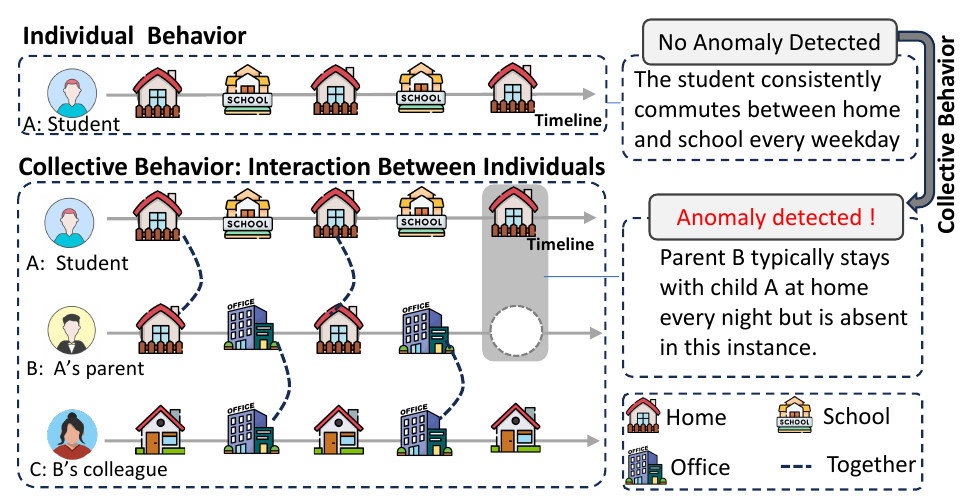}
    \vspace{-0.25in}
    \caption{Illustration of Individual and Collective Spatio-Temporal Behaviors. Modeling collective behaviors enables the detection of anomalies arising from people's interactions---referred to as collective anomalies---that can not be detected using individual behavior alone.}
\label{fig:collective_behavior_example}
    \vspace{-0.1in}
\end{figure}

 Detecting anomalies in human mobility is crucial for applications such as policy development \cite{adey2004surveillance,palm2013rights}, urban planning~\cite{barbosa2018human}, and public safety~\cite{meloni2011modeling,barbosa2018human,numosim}. Traditional anomaly detection methods primarily focus on individual movement patterns (e.g., a student typically stays at school on weekdays) and identify deviations from a person's routine behaviors (e.g., the student unexpectedly spends a weekday afternoon in a casino). However, in the real world, people often interact with others, forming collective behaviors. For example, as shown in ~\figref{fig:collective_behavior_example}, parent B typically stays with their child A at home at night or with their colleague C at the office during weekdays. These collective behaviors reveal anomalies driven by human interactions (termed collective anomalies), 
which are invisible at the individual level and have been largely ignored in previous research. This gap motivates our work: ``How can we effectively detect spatio-temporal anomalies based on \textit{collective} behaviors?'' 
% These group patterns are also relevant for anomaly detection but have been largely overlooked in previous literature.

%However, solving the above problem is non-trivial because of the following challenges:

% collective behavior plays an equally important role, and detecting anomalies within collective mobility patterns remains an open challenge. 

% if a student typically has lunch with a friend but unexpectedly eats alone one day,
\par Unlike individual anomalies, collective anomalies require modeling the spatiotemporal dependencies between individuals over time, adding an additional layer of complexity. For instance, in ~\figref{fig:collective_behavior_example}, student A typically stays at home with their parent, but unexpectedly is alone one day, which indicates a collective anomaly. Detecting such collective anomalies accurately presents three key challenges: (1) \textbf{Intricate Spatiotemporal Dependencies for Individual:} Even for a single individual, it is non-trivial to learn the behavior patterns from their event sequences that involve a mix of numerical and categorical spatiotemporal features (e.g., location, day of week, time of visit, stay duration,  etc.).
(2) \textbf{Collective Behavior Dependencies Modeling:} Capturing the relationships between different individuals and learning their collective behaviors require modeling relational dependencies in addition to the spatiotemporal ones. 
(3) \textbf{Collective Anomaly Detection}: Even after capturing these dependencies, it remains unclear how to effectively leverage them for collective anomaly detection.

As shown in~\tabref{tab:method_compare}, although extensive studies have been done on human mobility modeling and anomaly detection,  none of the existing approaches effectively addresses all three challenges discussed above. Specifically, human mobility modeling methods~\cite{2021CTLE,2023LightPath,DeepMove,want2024trans,MobTCast} focus on capturing spatiotemporal dependencies of a single individual, with limited or no consideration of collective behaviors. Similarly, trajectory anomaly detection~\cite{IBAT2011,ATROM,song2018anomalous,onlineGMVSAE2020,li2024difftad} methods are effective for identifying individual anomalies, but often fail to generalize to collective settings where interpersonal interactions are essential.  One might consider adapting (dynamic) graph-based AD models  \cite{rossi2020temporal,xu2020inductive,povstuvan2024learning}. However, those methods typically  model each incoming edge as an event, and thus flag individual edges as link anomalies (e.g. transactions). Moreover, they are not designed to incorporate spatial information, which is critical in human mobility contexts. In contrast, we  
%offline setting allows us to leverage both pre- and post-$t$ information, 
model collective interactions beyond individual edges, incorporating spatiotemporal context, and extend the task to also identify whether any unobserved link is anomalous. Previous works on collective behavior anomaly detection~\cite{ye2015discovering,rayana2015collective,van2015guilt,Shehn2023fraud, Haghighi2024fraud} primarily focus on group-level anomalies (e.g., collective opinion, group fraud reviews) without modeling spatio-temporal context. Moreover, their definition of collective behavior differs from ours: they focus on behavioral similarity within groups instead of co-occurrence in space and time used in our setting.
In summary, existing work does not address anomaly detection in the presence of collective behaviors.

% One might consider adapting graph-based models  \cite{rossi2020temporal,xu2020inductive,povstuvan2024learning}, which represent individuals as nodes and capture their interactions through links. However, link anomaly detection methods in continuous-time dynamic graphs typically aim to flag each incoming edge at time $t$ as anomalous, using only the history up to $t$. Moreover, these models are not designed to incorporate spatial information, which is critical in human mobility contexts. In contrast, our offline setting allows us to leverage both pre- and post-$t$ information, and we extend the task to also identify whether any unobserved link is anomalous. In summary, existing work does not address anomaly detection in the presence of collective behaviors.

%% However, in traditional link anomaly detection on continuous-time dynamic graphs, the goal is to flag each incoming edge at time $t$ as anomalous using only the history up to $t$. Also, those methods are not designed to model the spatial information as in our scenarios. 

% dynamic incoming

\begin{table}[!t]
% \begin{table}[htbp]
	    \centering
	    \caption{Comparison between our model and related works. (Abbr;  ISD: Intricate Spatiotemporal Dependencies for Individual, CBD: Collective Behavior Dependencies; CAD: Collective Anomaly Detection; AD: Anomaly Detection.}
        \vspace{-0.1in}
        % DR: Dynamic Relation, TAS: Temporal-order Asynchronous Sequence, MTF: Mixed-type, numeric\&categ,  features), MTS: Multivariate Time Series.
            % \vspace{-0.15in}
           % \renewcommand\arraystretch{1.5}
           \setlength\tabcolsep{1.5 pt}
    	\resizebox{1 \linewidth}{!}{
    		\begin{tabular}{lcccc}
    			\toprule
    			Methods & ISD & CBD & CAD \\
    			\midrule
    			  Human mobility modeling \cite{2021CTLE,2023LightPath,DeepMove,want2024trans,MobTCast}  &  \ding{52} &    &  \\
                    %Uncertainty learning in deep models \cite{kendall2017,Huang2018semantic,sensoy2018evidential,ye2024uncertaintyregularized,chanestimating}       &  & \ding{52} &  &  & \ding{52}\\
    			  Trajectory anomaly detection  \cite{IBAT2011,ATROM,song2018anomalous,onlineGMVSAE2020,li2024difftad} & \textbf{\ding{52}}   &  &    \\
                    %MTS anomaly detection \cite{2024DualTF,Feng2024HUE,zamanzadeh2024deep,wang2024revisiting}  &  &  & \textbf{\ding{52}}   \\

                    % Temporal graph models 
                    Graph-based AD models
                    \cite{rossi2020temporal,xu2020inductive,povstuvan2024learning}  &  & \textbf{\ding{52}} & \\

                    Collective Behavior AD~\cite{ye2015discovering,rayana2015collective,van2015guilt, Shehn2023fraud, Haghighi2024fraud} & & \textbf{\ding{52}} &  \textbf{\ding{52}}\\ % todo

    			\midrule
    			 \method (this paper)  & \ding{52} & \textbf{\ding{52}} & \textbf{\ding{52}}   \\
    			\bottomrule
    		\end{tabular}
    	}
	\label{tab:method_compare}
  \vspace{-2em}
\end{table}

To fill the research gap, we present \method, the first attempt to model collective behaviors for human mobility anomaly detection. We first formulate the problem as learning on Collective Event Sequences (CES) with an event graph. CES refers to multiple stay-event sequences of related individuals. The event graph captures the relationships between stay events, where nodes represent individual events, and edges (links) indicate co-occurrence in the same location. Then, \method introduces a Two-Stage Attention (TSA) mechanism to jointly model \textit{spatiotemporal} dependencies \textit{within} individual event sequences and \textit{relational} dependencies \textit{between} different individuals. The model is trained in an unsupervised manner to reconstruct both the attributes of randomly masked events (capturing individual behaviors) and their associated links (capturing collective behaviors). At inference, the event features and links are used to compute a carefully-designed  score for anomaly detection. The following summarizes the main contributions of this work:

% To fill the research gap, we present \method, the first attempt to model collective behaviors for human mobility anomaly detection. We first formulate the problem as learning on Collective Event Sequences (CES) with event-graph. CES refers to multiple stay-event sequences from socially or spatially related individuals. The event-graph captures interactions between stay events, where nodes correspond to individual events and edges (link) represent co-occurrences at the same location. 

% where the timing of events between individuals is not necessarily synchronized. 
%We employ an unsupervised learning approach based on the masked node and edge prediction task, where certain event nodes and their connections are randomly masked during training. The model is then trained to reconstruct both the event attributes through a node reconstruction loss and the relational dependencies using a link prediction loss, ensuring it effectively captures both individual behaviors and collective mobility patterns. % we propose \method, which captures both temporal dependencies within individual mobility sequences and relational dependencies between different individuals. To achieve this, we 

\begin{itemize}[leftmargin=*]
    % \item \textbf{Problem Formulation:} We present a new and more practical problem that introduces collective behaviors for human mobility anomaly detection, and 

     \item \textbf{Problem Formulation:} We present a new and more practical problem formulation for human mobility anomaly detection that explicitly incorporates collective behaviors. This setting enables learning from group-level mobility patterns and highlights their benefits in downstream tasks such as anomaly detection. 
    
    %We cast collective human mobility detection as unspervised learning on CES sequence with graph, which stores collective behaviors as multiple event sequences with temporal-order synchronicity, along with the event-graph to capture the co-occurrence relations.

    \item \textbf{Collective Behavior Modeling:} We introduce \method for human collective mobility modeling, bridging the gap between spatiotemporal and graph based models.
    It incorporates a two-stage attention mechanism and unsupervised pre-training to jointly learn individual mobility patterns as well as relational dependencies across individuals.

    \item \textbf{Collective Anomaly Detection:} Building on collective behavior modeling, \method proposes a tailored anomaly scoring function that integrates both individual and collective behavior to effectively detect different types of anomalies.
    
    %presents the first attempt for collective anomaly detection based on the anomaly detection function.

    \item \textbf{Effectiveness:} Extensive experiments on industry-scale data show that \method improves AUCROC  by 13\%-18\% and AUCPR 
    %\lmn{AUCROC and AUCPR -- please be consistent} 
    by 19\%-70\% over various anomaly detection baselines.
    
\end{itemize}

%while lacking the ability for collective anomaly.    Existing human mobility modeling methods~\cite{2021CTLE,2023LightPath,DeepMove,want2024trans,MobTCast} primarily focus on capturing spatiotemporal dependencies of one individual but lack explicit mechanisms for modeling collective behaviors. Meanwhile,  trajectory anomaly detection techniques are well-suited for analyzing individual movement trajectories but often fail to generalize to collective settings where interpersonal interactions matter. 

% To study the collective anomaly, we need to injest mutiple seqences with  

% support and evaluate real-world policy development and decision-making research related to mobility patterns and anomaly detection.

%collective anomaly in human event ad detection aims to find the anomaly group of the user by event sequences of multiple people.

%And to our best knowledge, there are few works on such topic. 

% The task is quite challenging since (1) it is already non-trival to learn the behavior patterns given a single users event sequences with 
% abdundant spatial-tempral features. (2) Simultanously consider multiple sequences of different people makes it more challenge to find the collective anomaly.

% To solve the above challenges, we first cast the problem with a novel formulation as learning on TAS with graph data. 
% Then we propose a xxx new to learn xxx,  the model is genarial and can be applied in many others filed.
% At last, to effectively detect the collective anomaly, we xxxx.

% To study the collective anomaly, we need to injest mutiple seqences with 

\section{Related Work}

% Methods in this domain broadly fall into two categories: traditional statistical models and deep learning techniques.
\textbf{Human Mobility Modeling.}   Statistical mobility modeling methods, such as Poisson/Hawkes processes \citep{Hawkes1971SpectraOS, daley2008point_processes, ogata1998space_time} and Markov Chains \citep{markov1, markov2, markov3, WhereNext}, depend on predefined functional assumptions to predict event arrival times or future locations. However, they often struggle to effectively model the complex spatiotemporal patterns inherent in human mobility. In contrast, deep learning approaches—particularly those built on RNNs \citep{Gao2017,Song2016,du2016recurrent} and Transformer architectures \citep{wan2021pre,DeepMove,MobTCast,Abideen2021taxitrans,wu2020transcrime,want2024trans}—have demonstrated strong performance in capturing intricate sequential transitions. Transformers, in particular, have become the prevailing choice due to their multi-head self-attention mechanism, which enables modeling interactions between all elements in a sequence. For example,  CTLE \cite{2021CTLE} incorporates context and time-awareness through masked pre-training of location embeddings. DeepMove~\cite{DeepMove} proposes an attentional recurrent neural network model that leverages historical trajectory patterns and temporal context to accurately predict the next location in human mobility sequences. Nevertheless, a common limitation of these methods is their failure to account for the collective mobility patterns. Such shortcomings motivate our work: a joint (spatiotemporal and relational) model of human collective mobility.
%\lmn{i am removing claims to be first, reviewers may object}
% the development of a collective mobility modeling framework to enhance human mobility modeling.

% \noindent \textbf{Human Mobility Anomaly Detection.} There is limited literature on human mobility anomaly detection; the most related work is trajectory anomaly detection, which aims to judge whether a trajectory is an anomaly given a sequence of GPS points. For example, IBAT \cite{IBAT2011} detects anomalies using how much the target trajectory can be isolated from other trajectories. GMVSAE \cite{onlineGMVSAE2020} uses a generative variational sequence autoencoder model that learns trajectory patterns with the Gaussian Mixture model and uses the probability of trajectory being generated as the anomaly score. ATROM \cite{ATROM} uses variational Bayesian methods and correlates trajectories with possible anomalous patterns with the probabilistic metric rule. However, since trajectory anomaly methods are primarily proposed for uniformly sampled GPS points, thus lack the ability to model the semantic complexity of human activities. Besides, such models cannot perform event-level anomaly detection when applied to our problem. 

\noindent \textbf{Human Mobility Anomaly Detection.} Research on human mobility anomaly detection remains scarce, with most existing studies focusing on trajectory anomaly detection, which determines whether an entire GPS trajectory is abnormal. For instance, IBAT \cite{IBAT2011} identifies anomalies based on how well a target trajectory is separated from others. GMVSAE \cite{onlineGMVSAE2020} employs a generative variational sequence autoencoder combined with a Gaussian Mixture Model to learn trajectory patterns, using the likelihood of trajectory generation as an anomaly score. Similarly, ATROM \cite{ATROM} leverages variational Bayesian techniques to associate trajectories with potential anomalous patterns via probabilistic rules. However, these methods are designed only for trajectory-level anomaly detection and therefore cannot be used for event-level anomaly detection. Moreover, they do not detect collective behavior anomalies, which is the core motivation behind our work.

\noindent \textbf{Collective Behavior Modeling.}
The concept of collective behavior varies across disciplines~\cite{ye2015discovering,rayana2015collective,van2015guilt,Shehn2023fraud}, but it generally refers to how individual actions give rise to emergent group-level patterns~\cite{bak2021stewardship}.
%Collective behavior can have different definitions under different disciplines. Typically, it refers to understanding how the actions of individuals can lead to emergent group-level phenomena \cite{bak2021stewardship}. 
Researchers have used diverse methodologies to study these dynamics, including agent-based models that simulate individual interactions \cite{larooij2025large}, and statistical physics approaches that seek to identify universal principles governing social dynamics \cite{castellano2009statistical,jensen2019politics}. Such models have been applied to domains like crowd behavior \cite{castellano2009statistical,jensen2019politics} and the emergence of social movements \cite{larooij2025large, marx1994collective}. In contrast to the classical definition, we define collective behavior in human mobility as the co-occurrence patterns between individuals. One possible approach for modeling such behavior is graph-based models \cite{rossi2020temporal,xu2020inductive,povstuvan2024learning}, where each individual is a node and interactions are encoded as edges. However, these typically do not model  sequences and thus the spatiotemporal dependencies across events and are not designed for collective anomaly detection. In summary, prior methods have largely overlooked anomalies arising from collective behaviors. Our work seeks to fill this gap by developing anomaly detection techniques that explicitly model these interactions.

% In summary, existing work does not address anomaly detection in the presence of collective behaviors. We aim to study how to achieve better anomaly detection with such collective behavior introduced.

%, studying biological swarming [14, 15], 

% is a multifaceted field that draws insights from various disciplines (e.g., sociology and psychology), to

%and exploring the formation and evolution of opinions within a population.[12, 16]
% network models that examine the influence of social structures [7, 3], 

% 5: From Mindless Masses to Small Groups: Conceptualizing Collective Behavior in Crowd Modeling
% 

% However, these methods are designed for uniformly sampled GPS data and therefore fall short in capturing the rich semantic context of human activities. Moreover, they are not equipped to support event-level and anomaly detection, which is essential for our application.

\section{Problem Formulation}

We formally define the related concepts and present the problem formulation for collective anomaly detection in human mobility.

\subsection{Human Mobility Anomaly Detection with Individual Event Sequence}

\noindent \textbf{Stay-point Event}. A stay-point event (simply event) is defined as a stationary point where an individual's GPS readings remain unchanged for at least five minutes. It represents the individual's daily activities, denoted by $e_i$ with spatiotemporal features $\bm x_i$:
\begin{equation*}
    {\bm x_i} = (x_i^{\rm st}, x_i^{\rm sd}, x_i^{\rm x}, x_i^{\rm y}, x_i^{\rm poi}, x_i^{\rm dow}) \;,
    \label{eq:input-features}
\end{equation*} 
where $\F_n = \{x_i^{\rm st}, x_i^{\rm sd}, x_i^{\rm x}, x_i^{\rm y}\}$ depicts the numerical feature set: $x_i^{\rm st} \in \R$ and  $x_i^{\rm sd} \in \R$ are the start time and stay duration of the event; $x_i^{\rm x}, x_i^{\rm y}$ are the two-dimensional coordinates depicting 
the latitude and longitude of the event's location.  $\F_c = \{x_i^{\rm poi}, x_i^{\rm dow}\}$ denotes the categorical feature set; $x_i^{\rm poi} \in \mathcal{P}$ is the Point-of-Interest (POI) such as office, store, etc. with $|\mathcal{P}|=\#$unique POIs  (see \tabref{tab:stat_test_data}), and  $x^{\rm dow} \in \mathcal{D}$ denotes the Day-of-Week (DOW) with $|\mathcal{D}|=7$. 

\noindent \textbf{Individual Event Sequence}.  An individual event sequence records all stay events of the individual in a given time period.  Let $\tau^{u}_{w} = \{e_1^{u}, \ldots, e_{L^{u}_w}^{u}\}$ be an individual $u$'s event sequence within given time window $w$ (e.g. three days)  sorted in chronological order, where $L_w^u$ is the sequence length. 
% where ${\bm x}_i^{u} \in \R^{F}$ is the feature vector of the $i$-th stay-point event with $F$ features and $L_w^u$ is the sequence length. 

%\par Previous efforts focus on human mobility anomaly detection with the individual event sequence (termed as individual anomaly detection for simplicity): 
\par Prior work on human mobility anomaly detection has primarily focused on detecting individual anomalies, using an individual's event sequence as input (termed as individual anomaly detection for simplicity):
Given an individual $u$'s event sequence over a $w$-day window ${\tau}_w^{u}$ and a target event $e \in {\tau}_w^{u}$, the goal is to design an anomaly score function $f$ that quantifies how anomalous $e$ is, i.e., how strongly it deviates from $u$'s typical behavior, formally:
\begin{equation}
    f_{AS}: (e; {\tau}_w^{u}) \mapsto {{\rm Anomaly ~Score}(e)} \;.
\end{equation}

\subsection{Human Mobility Anomaly Detection with Collective Event Sequences} \label{sec:problem-definition-ces}

% a father goes to school to pick up his child at 4 pm every day.
% , thus generating co-occur \& co-location events
Beyond individual events, a person $u$'s event may involve others due to social interactions. For instance,  $u$ usually attends morning meetings with his colleagues every Tuesday. To model such collective behavior, we introduce collective event sequences (CES) for the target individual $u$.

\noindent \textbf{Co-occurrence Relation of Two Events.} Let ${e}^{u_i}_k$ be the $k$-th event of person $u_i$, and similarly ${e}^{u_j}_l$ be the $l$-th event of person $u_j$. Two events ${e}^{u_i}_k$ and ${e}^{u_j}_l$ have a co-occurrence relation when their spatial distance is within a specified threshold and their temporal intervals overlap, that is,

\begin{tabular}{c}
 $ \qquad {\rm Distance} ((x_{u_i, k}^{{\rm x}}, x_{u_i, k}^{{\rm y}}), (x_{u_j, l}^{{\rm x}}, x_{u_j, l}^{{\rm y}})) < \delta \text{ and }$  \\
    $ \qquad [x_{u_i, k}^{{\rm st}}, x_{u_i, k}^{{\rm st}}+x_{u_i, k}^{{\rm sd}}] \cap [x_{u_j, l}^{{\rm st}}, x_{u_j, l}^{{\rm st}}+x_{u_j, l}^{{\rm sd}}] \neq \phi\;.
    $
\end{tabular}

\noindent
 where ${\rm Distance}(\cdot)$ calculates the Haversine distance~\cite{vanbrummelen2017heavenly} between two locations in geographical space. We define  individuals $u_i$ and $u_j$ to \textit{interact} if their events have a co-occurrence relationship. 

% who interact with $u$ within the time window $w$
\noindent \textbf{Collective Event Sequences}. Given the event sequence of a target person, ${\tau}^u_{w}$, its collective event sequences ${\mathcal{T}_w^{u}}$ are defined as a list of sequences $\{ {\tau}^{u_1}_{w}, \cdots, {\tau}^{u_m}_{w} \}$ belonging to $u$'s related individuals within the time window $w$. The related individuals are denoted as $\mathcal{R}^{u}_w = \{u_1, \cdots, u_m\}$.  In this paper, $\mathcal{R}^{u}_w$ contains two types of related individuals: $i$) \underline{C}o-\underline{O}ccurrence individuals (${CO}^{u}_w$) within the given time window, i.e, individuals who have at least one event that co-occurs with any event in $\tau^u_w$;
$ii)$ \underline{F}requently-\underline{M}eeting individuals ($FM^{u}$) within the given training period, i.e., individuals who interact with the target person at least twice with a total interaction duration exceeding 2 hours. Note that ${CO}^{u}_w$ is designed to capture all people who interact with $u$ in the specific time window, while $FM^{u}$ includes those with frequent interactions across the broader training period (which can extend beyond $w$). These two sets may partially overlap, but are not guaranteed to be the same. 
%For example, person $u^{\prime}$ who only  interact with $u$ once in the time window $w$, so that $u^{\prime} \in {CO}^{u}_w$ while  $u^{\prime} \notin {FM}^{u}$. 
For example, a person $u^{\prime}$ who interacts with $u$ only once during $w$ would belong to ${CO}^{u}_w$ but not to ${FM}^{u}$.
Overall, we define the final set of related individuals as the union of both: $\mathcal{R}^{u}_w = {CO}^{u}_w \cup {FM}^{u} $.

% whose activities overlap with (i.e., occur in the same place at the same time) the target person $u$ for at least one event in the given time window $w$;

\begin{figure}[!t]
    \centering
    \includegraphics[width=1 \linewidth]{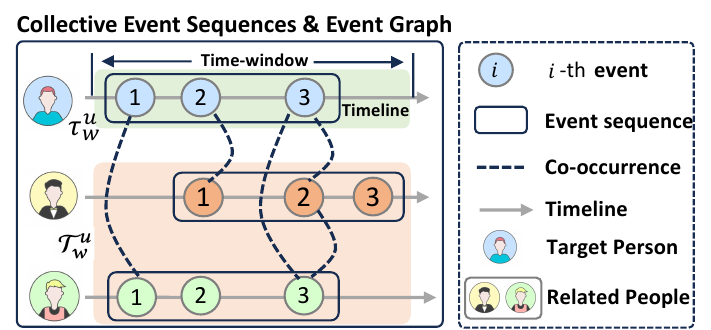}
    \vspace{-0.2in}
    \caption{Illustration of Collective Event Sequences (CES) and Event Graph. In this example, the target person has two related people in the given time window.}
    \label{fig:input}
    \vspace{-0.2in}
\end{figure}

\begin{figure*}[h]
    \centering
    \includegraphics[width=1\linewidth]{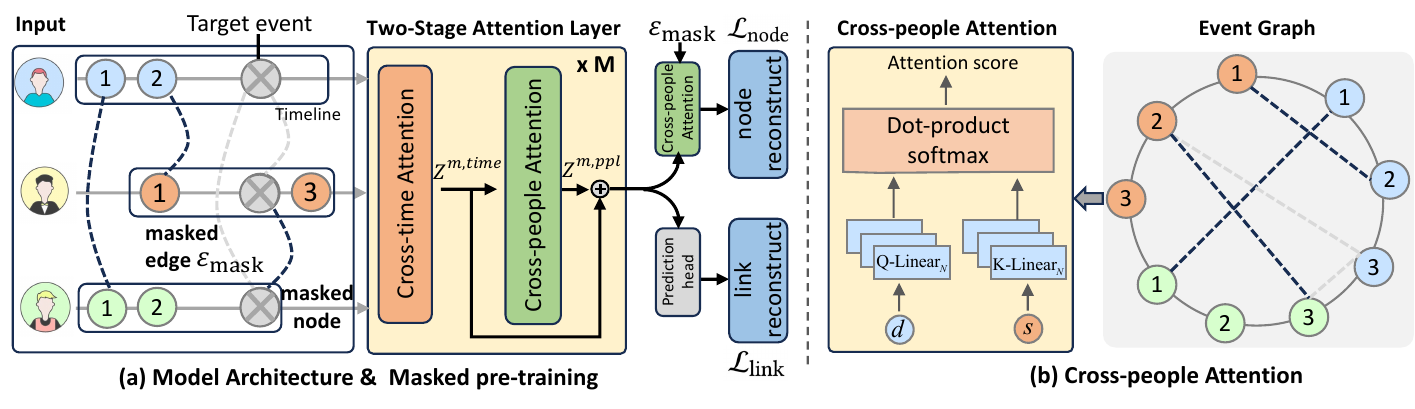}
    % \caption{Illustration of the input event sequence with co-locate \& co-occur graph.}
\vspace{-0.2in}
    \caption{\method model architecture,  which learns collective human mobility patterns in a self-supervised fashion via masked pretraining. \method employs a two-stage attention layer with (1) a \textit{cross-time} attention module to capture the \textit{spatiotemporal dependencies} in an event sequence of each single individual and (2) a \textit{cross-people} attention module based on the graph transformer, to capture the collective interactions or \textit{relational dependencies} between related individuals. }
    \label{fig:model}
\end{figure*}

\noindent \textbf{Event Graph}. The event graph for $u$ within the time window $w$ is defined as \( \G_{w}^{u} = (\V, \Edge) \), where the node set \( \V \) represents events from all sequences in \( \mathcal{T}_{w}^{u} \cup \{ {\tau^{u}_{w}} \} \), i.e. 
    $\V = \bigcup_{\tau \in \mathcal{T}_{w}^{u} \cup \{ {\tau^{u}_{w}}\}} \{ e | e \in \tau \}$. 
Edges \( \Edge \) represent the co-occurrence relationships between events from different sequences.
%, defined as pairs of events whose spatial distance is within a specified threshold (40 meters in this paper) and whose temporal intervals overlap. 
%Formally, an edge \( e_{(i,k), (j,l)} \in \Edge \) exists if the two events \( \bm{x}_k^{u_i} \) and \( \bm{x}_l^{u_j} \) (where \( i \neq j \)) have a co-occurrence relationship. 
~\figref{fig:input} shows an example of CES and the event graph.
{Notice that nodes in our graph depicts stay-events of individuals (in a sequence), and \textit{not} the individuals themselves as in many graph-based approaches, rendering it a more granular way of modeling interactions.}

% i.e.,  ${\rm Distance} ((x_k^{u_i, {\rm x}}, x_k^{u_i, {\rm y}}), (x_l^{u_j, {\rm x}}, x_l^{u_j, {\rm y}})) < 40$ and $[x_k^{u_i, {\rm st}}, x_k^{u_i, {\rm st}}+x_k^{u_i, {\rm sd}}] \cap [x_l^{u_j, {\rm st}}, x_l^{u_j, {\rm st}}+x_l^{u_j, {\rm sd}}] \neq \phi$.

% \noindent \textbf{Event-relation Graph}. The event-relation graph is defined as \( G = (V, E) \), where \( V \) represents events from all sequences in \( \mathcal{T}_{w}^{u} \cup \{ {\tau^{u}_{w}} \} \), and \( E \) denotes the edges that capture the correlations between these events. Specifically, the vertices \( V \) are constructed by pooling all events across the \( N \) sequences, i.e.,  
% \begin{equation}
%     V = \bigcup_{n=1}^N\big\{ \bm{x}_1^n, \bm{x}_2^n, \ldots, \bm{x}_{L_n}^n \big\}.
% \end{equation}
%The edges \( E \) in the graph represent the co-occur \& co-locate relationship between events from different sequences.  Formally, an edge \( e_{(i,k), (j,l)} \in E \) means that the two items \( \bm{x}_k^i \) and \( \bm{x}_l^j \) (where \( i \neq j \)) are correlated. We give an illustration of CES and event-relation graph in Figure~\ref{fig:input} by a specific example.

\noindent \textbf{Unsupervised Collective Human Mobility Anomaly Detection}. Given an individual $u$'s event sequence over a $w$-day window $\tau_w^u$, and their corresponding CES ${\mathcal T}_w^{u}$ with event graph $\G_w^u$, our goal is to learn an anomaly score function $f$ that quantifies how anomalous a target event $e \in {\tau}_w^{u}$ is. This function is designed to capture both the individual as well as collective behavior patterns, without relying on any labeled data, formally:
\begin{equation}
  f_{AS}: (e; {\tau}_w^{u}, {\mathcal T}_w^{u}, \G_w^u ) \mapsto {{\rm Anomaly ~Score}(e)}.
\end{equation}
%where the TAS is the event sequences of different individuals and the item-graph is defined as the co-located and co-occur graph between different events.

% i.e., how much it deviates from $u$'s typical (individual or collective) behaviors

 % \newpage

\section{Proposed Model: \method}

\figref{fig:model} presents the architecture of \method. The model first learns collective human mobility patterns in a self-supervised manner via masked pretraining, and then leverages the learned patterns for human mobility anomaly detection.
Unlike most prior work, \method explicitly models collective behavior through two key mechanisms: (1) \textbf{cross-time} attention to capture the \textit{spatiotemporal} dependencies within an individual's event sequence, and (2) \textbf{cross-people} attention, implemented via a graph transformer~\cite{dwivedi2020generalization}, to capture \textit{relational} patterns between related individuals (\S\ref{sec:collective_behavior_modeling}). The model is trained using masked pretraining, where the objective is to reconstruct both masked event attributes and co-occurrence links.
At inference time, the learned collective patterns are combined with the reconstruction error to compute the anomaly score (\S\ref{sec:collective_anomaly_detection}).

To facilitate the presentation, we pad the target person's sequence and all collective sequences to the maximum sequence length $L = {\rm max}\{L^u_w, \cdots, L^{u_\text{max}}_w\}$. The padded input is then represented as a tensor ${\bm X} \in \mathbb{R}^{N \times L \times F}$ where $N=\text{max}+1$ is the number of individual sequences, and $F$ is the number of event features.

% \blue{haomin: add ${\bm x}_i^{u} \in \R^{F}$ is the feature vector of the $i$-th stay-point event with $F$ features}

%Core idea: As shown in Figure~\ref{fig:tsa-layer}, the model consists of multiple Two-Stage Attention (TSA) layers inspired by \cite{zhang2023crossformer}. Let ${Z} \in \mathbb{R}^{T \times N \times D}$ be the embedding matrix after the embedding layer, where $D$ is the dimension of the event embedding. Recall that we use time step $t$ to index the $t$-th time segment and let $j \in \{1, \cdots, L\}$ index the $j$-th event. TSA (2-stage) layer takes $Z$'s as input and outputs next $Z$'s for all time steps $t$ and people $i$.
% \in  \{1, \cdots, 24\}

%  \begin{figure}[htbp]
%     \centering
%     \includegraphics[width=0.5\linewidth]{img/tsa_layer.pdf}
%     \caption{Illustration of the TSA layer.}
%     \label{fig:tsa-layer}
% \end{figure}

% \begin{figure*}[htbp]
%     \centering
%     \includegraphics[width=1\linewidth]{img/model2.pdf}
%     % \caption{Illustration of the input event sequence with co-locate \& co-occur graph.}
%     \caption{Model architecture,  which learns collective human mobility patterns in a self-supervised paradigm via masked pretraining. \method introduces two-stage attention layer with cross-time attention to capture the spatio-temporal dependence of one individual and cross-people attention module based on the graph transformer, to capture the collective behavior patterns between different individuals. }
%     \label{fig:model}
% \end{figure*}

\subsection{Collective Behavior Modeling} \label{sec:collective_behavior_modeling}

% \textbf{Feature Tokenizer.} 
% The input embedding layer converts the original input ${\bm X} \in \mathbb{R}^{N \times L \times F}$ into the embedding space ${\bm Z} \in \mathbb{R}^{N \times L \times D}$, where $F$ is the number of input features and $D$ is the embedding dimension. Specifically, we design different project layers for each feature to enrich the capacity of the model:  a numerical feature is projected by a linear transformation, while a categorical feature is projected by a learnable embedding layer. 

\subsubsection{Input Embedding Layer.} The input embedding layer transforms the raw input ${\bm X} \in \mathbb{R}^{N \times L \times F}$ into the embedding space ${\bm Z} \in \mathbb{R}^{N \times L \times D}$, where $F$ is the number of input features and $D$ is the embedding dimension.  Specifically,  each numerical feature $x^{(num)}_j$ is projected by a linear transformation with weight {${\bm W}^{(num)}_j \in \R^{F\times D}$} and bias ${\bm b}_j \in \R^{D}$. For categorical features, the embeddings are obtained via lookup from an embedding matrix ${\bm W}^{(cat)}_j \in \R^{C_{j} \times D}$, where $C_j$ is the total number of categories for feature $x_j^{(cat)}$, represented as a one-hot vector. 

In addition, we introduce three types of positional encodings $PE \in \R ^ {3 \times D}$ for each event to incorporate the temporal information: ($i$) Sequence positional encoding: encodes the event's position in the full sequence; ($ii$) Within-day positional encoding: captures the order of events within each day, delineating day boundaries; ($iii$) Day positional encoding: captures which day the event occurs on.
Each positional encoding is transformed to a non-learnable embedding vector with dimension $D$ following \cite{Transformer}. The final embedding $\z$ for an event is calculated by summing all its feature embeddings and positional embeddings, as follows.
\begin{alignat*}{2}
    &\z^{(num)}_j  =  x^{(num)}_j \cdot {\bm W}^{(num)}_j + {\bm b}^{(num)}_j  && \in \R^D \;, \\
    &\z^{(cat)}_j  = {x}^{(cat)}_j {\bm W}^{(cat)}_j             && \in \R^D \;, \\
    &\z            = \mathtt{sum} \left( \z^{(num)}_1,\ \ldots,\ \z^{(num)}_{k^{(num)}},\ \z^{(cat)}_1,\ \ldots,\ \z^{(cat)}_{k^{(cat)}}, PE \right)              && \in \R^{D} \;.
\end{alignat*}

% The feature embedding ${\bm e} \in \R^{F \times D}$ of one event is the concatenation of all numerical embeddings and categorical embeddings. 

% \par Recall that the original input is a sequence of sequence $\mathcal{T}$, we pad all sequences to the max number of sequences $L = {\rm max}\{ L_0, \cdots, L_M\}$, resulting the input tensor ${Z} \in \mathbb{R}^{N \times L \times D}$. Then $Z$ will be fed into several TSA layers, with the two attention mechanisms detailed as follows:

 \subsubsection{Two-stage Attention (TSA) Layer} Next, event embeddings are  fed into $M$ TSA layers to capture both individual and collective behavior patterns. Each layer contains cross-time attention followed by cross-people attention. Let $\Z^{m}$ be the embedding at the $m$-th layer, where the initial input is $\Z^{0}=\Z$, which is the output from the embedding layer.

\noindent \textbf{Cross-time Attention:} Cross-time attention is designed to capture spatiotemporal dependencies between events within a single person's sequence of events over time where the spatial information is represented by the event features. It applies the Multi-head Self-Attention (MSA) mechanism \cite{Transformer} along the time axis. Let $\Z_{i,:}^{m} \in \mathbb{R}^{L \times D}$ denote all event embeddings of person $i$ from the $m$-th TSA layer. The cross-time attention can be formulated as follows:
\begin{equation}
    \begin{aligned}
         {\widehat \Z}^{m,time}_{i,:} & = {\rm LayerNorm} (\Z_{i,:}^{m-1} + {\rm MSA}^{time} (\Z_{i,:}^{m-1}, \Z_{i,:}^{m-1}, \Z_{i,:}^{m-1})) \\
        {\Z}^{m,time}_{i,:} & = {\rm LayerNorm}({\widehat \Z}^{m,time}_{i,:} + {\rm MLP}({\widehat \Z}^{m,time}_{i,:})). \\
    \end{aligned}
    \label{eq:tsa}
\end{equation}
After cross-time attention,  embedding ${Z}^{m,time} \in \mathbb{R}^{N \times L \times D}$ captures the time dependencies between events for each person.

\noindent \textbf{Cross-people Attention:} 
In real-world settings, an individual's event embedding, e.g., the $k$-th event $\Z_{u,k}^m$ of person $u$, can be influenced not only by their own past or future events but also by the events of other individuals. A natural approach to model such dependencies is to aggregate information along the ``people axis''. For instance, one might consider collecting the $k$-th events from all individuals (i.e., $\Z_{:,k}^m$) and applying neural networks such as CNN~\citep{he2019stcnn} or GCN~\citep{zhao2022spatial}, as is common in spatiotemporal learning literature~\cite{yuspatio,guo2019attention,zhao2019t}. However, a prerequisite for this approach is that the $k$-th events across different individuals are aligned in time, which does not hold in our setting. Since input sequences are asynchronous in temporal axis, the $k$-th event in two sequences (e.g., $\tau_w^{u_i}$ and $\tau_w^{u_j}$) may correspond to different timestamps, i.e., $t(e_k^{u_i}) \neq t(e_k^{u_j})$, where $t(\cdot)$ denotes the start time  of an event. This misalignment renders position-based aggregation unreliable, as illustrated in ~\figref{fig:input}.

%In real-world scenarios, person $u$'s $k$-th event $\Z_{u, k}^m$ is not only correlated with other events of this individual but can also correlate with events of other individuals. To model such correlation, an intuitive idea is aggregating information along the person-axis by fetching all the $k$-th events of others (i.e., $\Z_{:, k}^m$), then applying neural networks such as CNN \cite{he2019stcnn} or GCN \cite{zhao2022spatial} aggregator as adopted in many spatio-temporal learning works. However, the aforementioned solution is not applicable since the input are temporal-order asynchronous sequences when the $k$-th item in any two sequences ($\tau_w^i$ and $\tau_w^j$) does not necessarily happen at the same time (as shown in Figure~\ref{fig:input}),  i.e., $t(x_k^i) \neq ~t(x_k^j),$ where $t(\cdot)$ is the time measurement of an item. 

% dwivedi2020generalization, hu2020heterogeneous

\par The above challenge motivates us to construct a graph $\G_{w}^{u}$ that contains all the events across different individuals, where edges represent co-occurrence relationships as defined in \S\ref{sec:problem-definition-ces} and illustrated in~\figref{fig:model}\blue{(b)}. Based on this graph,  we employ cross-people attention to better capture the dependencies between events from different individuals, inspired by the graph transformer \cite{dwivedi2020generalization}. Specifically, it is a destination-specific aggregation approach where each node only attends to its neighbor nodes as captured by $\G_{w}^{u}$. Let $(s,d)$ depict a a co-occurrence relationship between source node $s$ and destination node $d$, and $\z_d^m = \Z^{m,time}[d]$ be the embedding of node $d$ after cross-time attention. Then, cross-people attention for the destination node $d$ is given as
\begin{equation}
    \begin{aligned}
    \widehat{\z}_d^m &= \z_d^m \|_{h=1}^H\left(\sum_{s \in \mathcal{N}(d)} w_{sd}^{m,h}~{\bm V}^{m,h}~\z_s^m\right) \\
    w_{s d}^{m,h} & =\operatorname{softmax}\left(\frac{{\bm Q}^{m,h} {\z}_d^m \cdot {\bm K}^{m,h} \z_s^m}{\sqrt{D}}\right) \;, \\
    % \operatorname{Attention}_{G T}(s, d) & =\underset{\forall s \in {\mathcal N}(d)}{\operatorname{Softmax}}\left(\|_{m \in[1, h]} {head}^{m}(s, d)\right) \\
    % { head }^m(s, d) & =\left(K^m(s)  Q^m(d)^T\right) \cdot \frac{1}{\sqrt{D}} \\
    % K_N^m(s) & =\text{K-Linear}^m_{N} \left(Z^{time}[s]\right) \\
    % Q^m(d) & =\text { Q-Linear }_{N}^m\left(Z^{time}[d]\right),
\end{aligned}
\end{equation}
where $D$ is the dimensionality of the embedding, and $H$ is the number of attention heads. For each node $d$, ${\mathcal N}(d)$ denotes its set of neighbors (i.e. related individuals). ${\bm Q}^{m,h}, {\bm K}^{m,h}, {\bm V}^{m,h} \in \mathbb{R}^{D \times D/H}$ are learnable matrices to project the node embeddings into query, key, and value vectors, respectively. After cross-people attention, we obtain the embedding $\Z^{m,ppl} \in \mathbb{R}^{N \times L \times D}$ which captures the relational dependencies across different people.

Finally, we aggregate the cross-time and the cross-people embeddings for each event to obtain the final output of the $m$-th TSA layer, that is, ${\Z}_{}^{m} = \Z^{m,time}_{} + \Z^{m,ppl}_{}$.

% ${\Z}_{i,j}^{m} = \Z^{m,time}_{i,j} + \Z^{m,ppl}_{i,j}$.

\subsubsection{Unsupervised Pre-training.} To train the model, we adopt the masked training strategy. As shown in~\figref{fig:model}\blue{(a)}, we first randomly mask a subset of events (called target events) from the target individual at a given ratio (set to 0.05). Then, for each target event $d$, we treat it as the destination node and mask all incoming links (edges) and source nodes. The model is trained to reconstruct all the features of the masked events and the masked links based on the input. Each feature $f$ is associated with a prediction head ${\rm Dec}_f$ (i.e, a linear projection), which decodes the output embedding $\Z^{M}$ after the collective behavior modeling into predictions.

% mask all features of a destination node and all incoming links (edges) of the destination node, and train the model to reconstruct the original features and links based on the input. Each feature $f$ is associated with a prediction head ${\rm Dec}_f$ (i.e, a linear projection), which projects the output embedding $\Z$ after collectvie behavior modeling into the predictions.

%where decoding the numerical features is considered as a regression while decoding the discrete feature is considered as a classification task. 

\par \noindent \textbf{Node Reconstruction Loss}. For each masked node, we decode its features via two types of tasks based on feature type. For numerical features, reconstruction is treated as a regression task, with loss
\begin{equation}
    \mathcal{L}_{\text {num}}= \sum_{f \in \F_n}\left\|y_f - {\widehat y}_f\right\|_2^2,
    \label{eq:loss_num}
\end{equation}
where ${\widehat y}_f$ and $y_f$ respectively refer to the prediction and the true value of the feature, and $||\cdot||^2$ refers to $\rm L_2$-norm. For categorical features, decoding is considered a classification task, where we calculate the cross-entropy loss
\begin{equation}
    \mathcal{L}_{\rm cls} = - \sum_{f \in \F_c} \sum_{i=1}^{C_f} y_i^f\log(\widehat{y}_i^f),
    \label{eq:loss_class}
\end{equation}
where $y_i$ is the one-hot encoded ground truth value, $\widehat{y}_i$ is the predicted probability (obtained via softmax) of value $i$, and $C_f$ is the number of unique values for feature $f$. Overall, the node reconstruction loss is the sum of these two terms. 
\begin{equation}
    \mathcal{L}_{\rm node}={\mathcal{L}_{\rm num}+{\mathcal{L}_{\rm cls}}} \;.
    \label{eq:node-loss}
\end{equation}
% $\mathcal{L}_{\rm node}={\mathcal{L}_{\rm num}+{\mathcal{L}_{\rm cls}}}$.

% Link reconstruction aims to model collective behavior patterns, specifically the interactions between individuals. To learn such patterns,

\par
\begin{figure}[htbp]
    \centering
     \vspace{-0.1in}\includegraphics[width=1\linewidth]{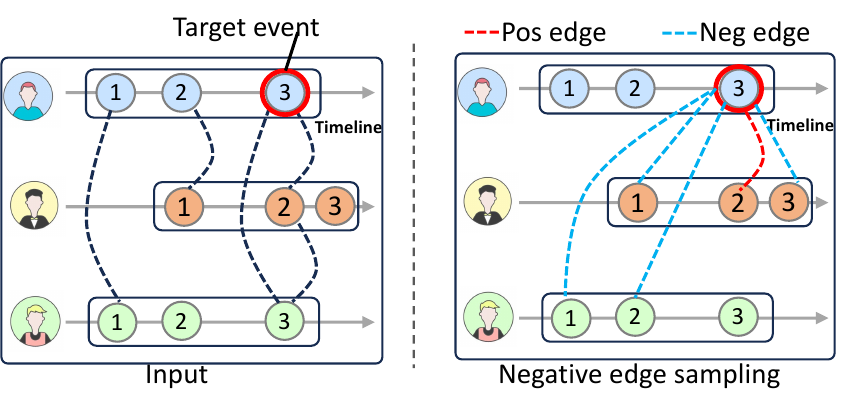}
    \vspace{-0.3in}
    \caption{Negative Edge Sampling. For each positive edge $(s,d)$, we sample a fixed number of negative edges ${\rm Neg}(s,d)=\{(s^{\prime},d) | {\rm  ~where~} s^{\prime}\in \{ {\mathcal V} \setminus  {{\mathcal N}(d)}
 \} \}$ uniformly at random.}
 %\vspace{-0.1in}
    \label{fig:neg_edge}
\end{figure}

 \noindent \textbf{Link Reconstruction Loss.} For link reconstruction, we mask all incoming edges to a target event $d$, treating each masked edge as a positive edge. For each positive edge $(s, d)$, we construct a corresponding set of negative edges  ${\rm Neg}(s, d)$ (see ~\figref{fig:neg_edge}) and train the model to distinguish positive from negative edges using the contrastive loss, i.e. 
\begin{equation}
    \resizebox{1 \linewidth}{!}{
        \begin{math}
        % \mathcal{L}_{\text{link}} = - \frac{1}{|\mathcal{E}|} \sum_{(s, d) \in {\mathcal E}} \Bigg\{ \log \left[ \frac{e^{\text{sim}(s, d)}}{e^{\text{sim}(s,d)} + \sum_{(s', d) \in \text{Neg}(s, d)} e^{\text{sim}(s', d)}} \right] \Bigg\},
         \mathcal{L}_{\text{link}} = - \frac{1}{|\mathcal{N}(d)|} \sum_{s \in \mathcal{N}(d)} \Bigg\{ \log \left[ \frac{{\rm \exp}^{\text{sim}(s, d)}}{{\rm \exp}^{\text{sim}(s,d)} + \sum_{(s', d) \in \text{Neg}(s, d)} {\rm\exp}^{\text{sim}(s', d)}} \right] \Bigg\},
         \end{math}
        }
    \label{eq:loss_link}
\end{equation}
where $\mathcal{N}(d)$ denotes $d$'s neighbors in the graph. The similarity between two events $s$ and $d$ is measured by $\mathrm{sim}(s, d)$, based on the cosine similarity between their embeddings $\mathbf{z}_s$ and $\mathbf{z}_d$. 
%$e$ is the natural logarithm base (a scalar $\approx 2.71828$).
%For a positive edge $(s, d)$, ${\rm Neg}(s, d)$ is its corresponding set of negative edges.

%${\rm sim}(s, d)$ is a similarity function (cosine similarity adopted in this paper) between the two embeddings $\z_s, \z_d$. 

%\vspace{-0.1in}
\floatname{algorithm}{Algorithm}  
\renewcommand{\algorithmicrequire}{\textbf{Input:}}  
\renewcommand{\algorithmicensure}{\textbf{Output:}} 
\begin{algorithm}[htbp]
   \caption{\method \textbf{Model Pre-training} }
   \small
\begin{algorithmic}[1]
   
   % \Require Target person sequence $\tau^u_w$, collective event sequence $\mathcal{T}_w^u$ and event-relation graph $\mathcal{G}^u_w$.

   \Require Training dataset $\mathcal{D}$, time window $w$.
   
   \Ensure Pre-trained model $\mathcal{M}$

   \Repeat

   \State Draw training sample ($\tau^u_w$,$\mathcal{T}_w^u$,$\mathcal{G}^u_w$)$\sim {\mathcal{D}}$

   \State Randomly mask target nodes in $\tau^u_w$ and their associated neighbor nodes and links in $\mathcal{G}^u_w$;
   
   \State Construct input $\bm X \in \mathbb{R}^{N \times L \times F}$;

   \State Init embedding $\bm Z^{0} \in \mathbb{R}^{N \times L \times D}$: Input-Embedding-Layer($\bm X$);

    \For{$m=1$ {\bfseries to} $M$}
        \State ${\Z}^{m,time}$ = CrossTimeAtt(${\Z}^{m-1}$)  $\blacktriangleleft$ $O(N \cdot L^2 \cdot D)$

        \State ${\Z}^{m,ppl}$ = CrossPeopleAtt(${\Z}^{m,time}, \G^u_w$) $\blacktriangleleft$ $O(|\V| \cdot D^2 + |\Edge|\cdot D)$

        \State ${\Z}^{m}$ = ${\Z}^{m, time} + {\Z}^{m, ppl}$

    \EndFor

    \State Calculate ${\mathcal L}_{\rm node}$ according to ~\eqnref{eq:node-loss};

    \State Calculate ${\mathcal L}_{\rm link}$ according to ~\eqnref{eq:loss_link};

   \State Calculate ${\mathcal L}_{\rm total}$ according to ~\eqnref{eq:loss_total}.

   \Until{converged}
\end{algorithmic}
\label{alg:training}
\end{algorithm}
\setlength{\textfloatsep}{0.1in}

\par \noindent \textbf{Negative Edge Sampling.} 
\figref{fig:neg_edge} illustrates the procedure for generating negative edges corresponding to a given positive edge. Given that $d$ is the target event, all the non-neighbor events of $d$ can be represented by $\{ {\mathcal V} \setminus  {{\mathcal N}(d)}
 \}$. Then, for each target edge $(s,d)$, we sample a fixed number of negative edges ${\rm Neg}(s,d)=\{(s^{\prime},d) | {\rm  ~s.t.~} s^{\prime}\in \{ {\mathcal V} \setminus  {{\mathcal N}(d)}
 \} \}$ as the negative edge set.

\par Overall, the total loss is the weighted sum of the node reconstruction  and link prediction losses, that is, 
\begin{equation}
    \mathcal{L}_{\rm total} = \mathcal{L}_{\rm node} + \lambda \mathcal{L}_{link},
    \label{eq:loss_total}
\end{equation}
where $\lambda$ (set to 0.01) is the hyperparameter to balance the scale of the two losses. Importantly, as shown in ~\figref{fig:model}\blue{(a)}, the two loss components are trained under different input conditions. Formally, let $\mathcal S$ be the input instance after masking, and ${\mathcal{V}_{\rm mask}}$ and ${\mathcal{E}_{\rm mask}}$ denote the nodes and links being masked. Then, the link prediction is  $p( {{\mathcal{\widehat E}}_{\rm mask} | {\mathcal S}})$, while the node reconstruction is  $p({\mathcal{\widehat V}_{\rm mask}} | {\mathcal{E}_{\rm mask}, {\mathcal S})}$. For node reconstruction, the masked links ${\mathcal{ E}}_{\rm mask}$ are added back after the TSA layer, enabling cross-people attention to better reconstruct the masked node features.

% takes in the ground-truth links (i.e., ${\mathcal{ E}}_{\rm mask}$) after TSA layer then followed by a 
% to better reconstruct the masked node features.

%For link reconstruction, we use the masked node and links as input, while for node reconstruction, we utilize the observed links to better reconstruct the masked node features. Formally, let ${\mathcal{V}_{\rm mask}}$ and ${\mathcal{E}_{\rm mask}}$ denote the masked nodes and links, and $\mathcal S$ be the input instance after masking
%then the link prediction is  $p( {{\mathcal{\widehat E}}_{\rm mask} | {\mathcal S}})$, while the node reconstruction is  $p({\mathcal{\widehat V}_{\rm mask}} | {\mathcal{E}_{\rm mask}, {\mathcal S})}$.

\subsection{Collective Anomaly Detection} 
\label{sec:collective_anomaly_detection}

Up to this point, we have learned the individual and collective mobility patterns. This raises two key questions: While prior methods typically rely solely on individual mobility patterns, how can we effectively incorporate collective behavior into anomaly scoring? Furthermore, what types of collective anomalies should be targeted, and how can they be detected at inference time using the pre-trained model?
% which raises critical questions: While previous methods often rely solely on individual mobility patterns, how can we effectively incorporate the collective mobility patterns into anomaly scoring?  Moreover, what types of collective anomalies should we detect and how can we detect them at the inference stage via the pre-trained model? 
In this section, we address these questions by defining different anomaly types and the solution to detect them.

As in many previous works, a common starting point for anomaly detection is to use the reconstruction error of node features.  For numerical features, it is defined as  ${PE}_f =  | y_f - \widehat{y}_f |$, where \(y_f\) is the observed value, \(\widehat{y}_f\) is the predicted value. For categorical features, it is ${PE}_f =  1 - \widehat{y}_c^f$, where $\widehat{y}_c^f$ is the predicted probability of the observed class of that feature. Overall, the node reconstruction error reflects how much the observation deviates from the model's learned individual behavior patterns.  The corresponding anomaly score\footnote{We apply percentile transform \cite{percentile} to each term before aggregation to make them comparable in scale.}  for a given event is defined as:
% Overall, the node reconstruction error can be considered as evaluating the deviation between the model learned individual patterns and the observation, the anomaly score function is formulated as follows:
\begin{equation} 
f_\text{AS}(e)_{\rm node} =  \max_{f \in \mathcal{F}_n \cup \mathcal{F}_c} \big\{PE_f \big\}.
\label{eq:ad-node-reconstruction}
\end{equation}

\begin{figure}[!t]
    \centering
    \includegraphics[width=0.9 \linewidth]{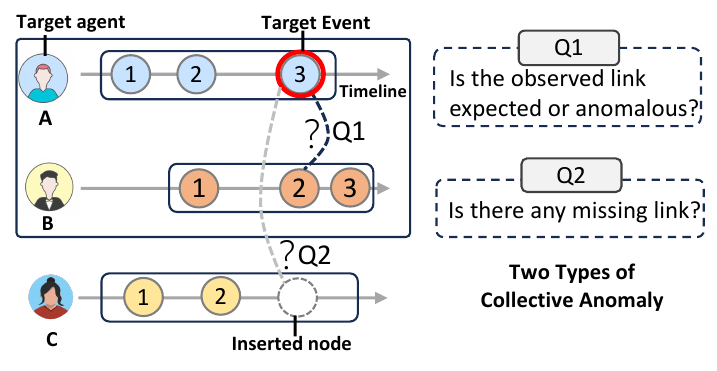}
    \vspace{-0.15in}
    \caption{Illustration of two collective anomaly types. Obs: In the target event A.3,  the target person A has lunch with B. We then ask: Q1: Is the A-B connection unusual? Q2: Is there someone else, e.g. C, who should be with A but is missing? }
   % \vspace{-0.1in}
    \label{fig:collective_ad}
\end{figure}

In addition to identifying individual anomalies, we incorporate collective behavior patterns to detect collective anomalies. Specifically, for a given target event, we aim to answer two key questions:
% Instead of detecting the individual anomalies, we also take the collective patterns into consideration.  Specifically, given a target event, we seek to answer two key questions to detect the collective anomaly: 
(1) Should the target individual be connected to their observed neighbors? In other words, \textit{is the observed link expected or anomalous}? (2) Are there any individuals with whom the target individual should be connected but are missing from their observed neighbors? In other words, \textit{is there any missing link?}  ~\figref{fig:collective_ad} illustrates a simplified example. Suppose the target person A has lunch with person B (event A.3). We ask:
(1) Is the observed co-occurrence between A and B unusual? If so, then event A.3 should be flagged as an anomaly since a stranger B appeared unexpectedly.  (2) Is there someone (e.g., person C) who typically joins A for lunch but is absent this time? If so, then event A.3 should be considered anomalous due to a missing link. To capture these two types of collective anomalies, we propose separate anomaly scores corresponding to unexpected links and missing links.

\noindent \textbf{Detecting unexpected co-occurrence anomaly}: If the model assigns a low probability to an observed link, it suggests that the link is unexpected or anomalous. Since a target event may have multiple neighbors, we define the anomaly score as the maximum deviation across all observed neighbors:
\begin{equation} 
f_{\text{AS}}(e)_{\rm obs\text{-}link} = \max_{s \in {\mathcal{N}(e)}}\{1 - S_{\rm link}(s, e)\}.
\label{eq:ad_score_obs_link}
\end{equation}
$f_\text{AS}(e)_{\rm obs\text{-}link}$ is designed to detect the observed but unexpected links, thus helping to detect the unexpected occurrence anomaly.

%we examine whether they should exist by the following anomaly score:

% % we classified the collective anomalies into two types related to the link: (1) should-not-exit link; (2) missing link;

\noindent \textbf{Detecting absence anomaly:} We consider unobserved links, i.e., individuals who are not present in the event but are expected to be.
%For unobserved links, a higher chance (given by the model) of a non-existing link means more anomalous, thus leading to the following anomaly score:
If the model assigns a high likelihood to a link that does not exist in the observation, this suggests an anomalous absence. The corresponding anomaly score is defined as:
\begin{equation} 
f_\text{AS}(e)_{\rm not\text{-}obs\text{-}link} = \max_{s \notin {\mathcal{N}(e)}}\{S_{\rm link}(s, e)\}.
\label{eq:ad_score_not_obs_link}
\end{equation}
% A challenge here is that the above anomaly score is calculated based on the embeddings of two nodes, while in the unobserved case, there are no nodes for us to compute the score. 
% A key challenge in this setting is that unobserved links correspond to missing events, i.e., there is no corresponding node embedding for the absent interaction. 
A key challenge in this setting is that unobserved links correspond to missing events, i.e., there is no corresponding node embedding to compute the above score. 
To address this challenge, we introduce a “ghost node” for each potential missing neighbor, representing an event that would have occurred during the time range of the target event but was not observed. These ghost nodes allow us to compute link scores even in the absence of explicit event data, as illustrated in~\figref{fig:collective_ad}.
% To address this challenge, as shown in Figure~\ref{fig:collective_ad}, we insert a ``ghost node'' where there is no event observed during the time range of the target event. 
Overall, the anomaly score is defined as the combination of the above three scores:
\begin{equation} 
f_\text{AS}(e) = \max\{f_\text{AS}(e)_{\rm node}, f_\text{AS}(e)_{\rm obs\text{-}link}, f_\text{AS}(e)_{\rm not\text{-}obs\text{-}link} \}. 
\label{eq:ad-score-node-link}
\end{equation}

%$|{\rm Nei}(e)|=0$ means that the event does not have a neighbor event, in that case, we do not consider its link anomaly score, and thus set $AS(e)_{link}=-inf$

\floatname{algorithm}{Algorithm}  
\renewcommand{\algorithmicrequire}{\textbf{Input:}}  
\renewcommand{\algorithmicensure}{\textbf{Output:}} 
\begin{algorithm}[htbp]
   \caption{\method: \textbf{Collective Behavior Anomaly Detection}}
   \small
\begin{algorithmic}[1]

   \Require Target event $e$, Target person sequence $\tau^u_w$, collective event sequence $\mathcal{T}_w^u$, pre-trained model $\mathcal{M}$.
   %event sequence of the target user $\tau_w^u$.
   
   \Ensure Anomaly score of $e$

   \State Add ``ghost node'' for sequences in $\mathcal{T}_w^u$, and construct event graph $\mathcal{G}^u_w$;
   \State Mask the target event $e$ in $\tau^u_w$ and its associated neighbor nodes $\mathcal{N}(e)$ and links in $\mathcal{G}^u_w$;

   \State Get model prediction $\Z, {\widehat Y}={\mathcal{M}(\tau^u_w, \mathcal{T}_w^u, \mathcal{G}^u_w)}$; $\Z$ for node embedding, ${\widehat Y}$ for reconstructed target event features;

   \State Calculate $f_{\text{AS}}(e)_{\rm node}$ in ~\eqnref{eq:ad-node-reconstruction} based on ${\widehat Y}$;

   \State Calculate $f_{\text{AS}}(e)_{\rm obs-link}$ in ~\eqnref{eq:ad_score_obs_link} and $f_{\text{AS}}(e)_{\rm not-obs-link}$ in ~\eqnref{eq:ad_score_not_obs_link} based on $\Z$;

   \State Return final anomaly score in \eqnref{eq:ad-score-node-link}.
   
\end{algorithmic}
\label{alg:ad}
\end{algorithm}
 \setlength{\textfloatsep}{0.1in}

% \begin{algorithm}[htbp]
%     \caption{Anomaly Detection of \method}
%     \small
%     \begin{algorithmic}[1]
%     \Require Historical graph signal $\xh$,  graph $\G$, trained denoising function $\be_{\theta}$
%     \Ensure Future forecasting $\xp{}$

%     \State Construct $\xall_{\rm msk}$ according to $\xh$

%     \State Sample $\be \sim {\mathcal{N}}(\mathbf{0}, \mathbf{I})$ where $\be$'s dimension corresponds to $\xall_{\rm msk}$
%     \For{$n=N$ {\bfseries to} $1$}
%         %\STATE Sample $\xall{n-1}$ using Eq.~(\ref{eq:condition_reverse_process}) and Eq.~(\ref{eq:condition_p_sample}) by taking $\xh$ and $G$ as condition
%         \State Sample $\xall_{n-1}$ using Eq.~(\ref{eq:masked_condition_reverse_process}) by taking $\xall_{\rm msk}$ and $\G$ as condition
%     \EndFor
%    \State Take out the forecast target in $\xall_{0}$, i.e., $\xp{}$
%     \State Return $\xp{}$
%     \end{algorithmic}
%     \label{alg:samping}
  
% \end{algorithm}

\noindent \textbf{{Complexity Analysis}}:  The pre-training of \method is outlined in Algorithm~\ref{alg:training} and the anomaly detection process is described in Algorithm~\ref{alg:ad}. Given collective event sequences with $N$ individuals and a maximum sequence length of $L$, where each event is represented as a $D$-dimensional hidden vector, the time complexity of the cross-time attention is $O(N \cdot L^2 \cdot D)$. The cross-people attention is only computed between a node and its neighbors (not all node pairs), resulting in a complexity of $O(|\V| \cdot D^2 + |\Edge|\cdot D)$. Here, $O(|\V| \cdot D^2)$ is the complexity for the linear transformation of node embeddings with a $D \times D$ projection matrix, where $|\V|$ is the number of nodes in the graph (with a maximum value of $N \cdot L$). The term $O(|\Edge|\cdot D)$ accounts for the cross-people attention, where $|\Edge|$ is the number of edges in the graph. Overall, the total complexity of the model is $O(N \cdot L^2 \cdot D + N \cdot L \cdot D^2 + |\Edge|\cdot D)$. This analysis shows that the model scales linearly with the number of sequences $N$ and quadratically with the maximum sequence length $L$.

% The matrix-vector multiplication xW requires: \mathcal{O}(d \cdot d{\prime}), when d{\prime} = d, that would be O(d^2)

%The model takes approximately 3 minutes to train one epoch (411k samples of batch size 128) and performs inference at an average speed of 0.04 seconds per sample for $T=50$ and $w=3$ on one GPU (Nvidia RTX A6000). Thus, it leads to 25 QPS (query per second) for batch size 1 for the deployed system, which can meet the speed needs of numerous real-world applications.

% \begin{equation} 
% \text{AS}(e)_{\rm link} = \max_{s \in {\rm Nei(e)}}\{1 - S_{\rm link}(e, s)\} 
% \end{equation}

% \begin{equation}
% \text{AS}(e)_{\rm link} = 
% \begin{cases}
% \bm{v}^T \tanh\left( \bm{W}_1 e_i + \bm{W}_2 h_j \right), & \text{if } i \ne \pi_{j'} \quad \forall j' < j \\
% -\infty, & \text{otherwise}
% \end{cases}
% \label{eq7}
% \end{equation}

\section{Experiments}

We conduct extensive experiments on two industry-scale datasets to answer the following questions: \textbf{Q1: Anomaly detection:} How does \method perform in anomaly detection compared to existing methods? What type of collective anomalies can \method aid detect? \textbf{Q2: Link prediction}: What is \method's performance in predicting masked edges?  \textbf{Q3: Ablation study:} How do different components of \method contribute to its performance? 
% \textbf{Q4: Scalability:} How does \method scale with respect to runtime and memory?  
\textbf{Q4: Case studies:} How can \method help us interpret the detected anomalies?
\textbf{Q5: Scalability:} Can we train \method efficiently on industry-scale datasets? 

% \begin{itemize}[leftmargin=*]
%     \item \textbf{Q1: Link Prediction}: What is the \method's performance in predicting masked edges?
%     \item \textbf{Q2: Anomaly detection:} How does \method perform in collective anomaly detection compared to previous anomaly detection?
%     \item \textbf{Q3: Ablation study:} How does different components contribute to \method's performance?

%     \item \textbf{Q4: Scalability:} How can \method scale up with respect to time and memory?
    
%     \item \textbf{Q5: Case studies} How does \method help us interpret the detected collective anomalies?

% \end{itemize}

\subsection{Experiment Setting}

% Novateur
% \textbf{Dataset}. 
\subsubsection{Datasets}
We evaluate different methods on two industry‑scale datasets — \trialfourone, \trialfourtwo — each containing tens of thousands of agents and millions of events that are simulated (1) in real‑world cities (2) by domain experts to guarantee that the data is both realistic and well mimics human behavior. Each dataset is divided into two disjoint activity periods: The training period consists exclusively of normal activities, allowing the model to learn typical behavioral patterns. The testing period includes a small fraction of injected anomalous activities to assess detection performance. 
%\trialfourone and \trialfourtwo are two private datasets where anomalies in both are manually injected. 
% following our defined strategies
% To ensure fairness and robustness, anomalies in \trialfourtwo are injected by independent “red teams” at (company name blinded). This setup guarantees that $i$) the anomalies are realistic, and $ii$) \textit{our modeling is agnostic to the way anomalies were introduced}. 
Detailed statistics are given in \appref{sec:data}, Table ~\ref{tab:stat_test_data}. In summary,
\begin{itemize}[leftmargin=*]
    \item \trialfourone is simulated in Tokyo with two-month mobility data for 20K agents, of which 2415 (12.1\%) are anomalous, and 3.47 million events, with 2595 (0.07\%) labeled as anomalous. 
    % \item \trialfourtwo is also simulated in Tokyo, containing 10K agents, of which 1599 (16\%) are anomalous, and 1.91 million events, with 38219 (1.9\%) labeled as anomalous. The anomalies are injected by the data simulation company.
    \item \trialfourtwo is also simulated in Tokyo, containing 20K agents, of which 3379 (17\%) are anomalous, and 3.51 million events, with 3635 (0.1\%) labeled as anomalous. 
\end{itemize}
\textit{Why Simulated Data?} There are several reasons for conducting experiments on simulated datasets: (1) First and most importantly, to the best of our knowledge, there is currently no human activity data available at the industry scale with collective behaviors and anomaly labels. While many real-world human trajectory datasets exist, such as GeoLife \cite{zheng2009mining}, Gowalla\footnote{https://snap.stanford.edu/data/loc-gowalla.html}, Foursquare\footnote{https://sites.google.com/site/yangdingqi/home/foursquare-dataset}, LaDe~\cite{wu2024lade}, and Instagram~\cite{chang2018content-instagram}, none of them include anomaly labels. Conversely, \NUMOSIM~\cite{numosim} provides labels for individual anomalies but lacks representations of collective behaviors. (2) Collecting real human mobility data and manually labeling anomalies are both time-consuming and resource-intensive. They often require domain expertise and careful contextual understanding, making them impractical at scale. (3) Simulated data enables us to inject various types of realistic anomalies systematically. This allows for a comprehensive evaluation of the model's ability to detect diverse anomaly types. 

% which originally contain no anomalies, to facilitate evaluation. 
\textit{Anomaly injection.} We inject 3 types of collective anomalies into \trialfourone and \trialfourtwo: (1) \textbf{Unexpected Occurrence:} We randomly select an event involving multiple individuals (i.e., a co-occurrence) and then choose another individual who did not participate in that event and was not at that location during the corresponding time window. We modify this individual's event by changing its location and POI to match the selected event. This creates an unexpected co-occurrence, as the individual was never expected to be part of that group event. (2) \textbf{Absence:} We randomly select a co-occurrence event and then randomly remove one of the participating individuals by altering their event's location (latitude and longitude) and POI type to a different place. This simulates an absence anomaly, where the individual is expected to attend the event but is instead missing. (3) \textbf{Synthetic Coordination:} We first identify a group of individuals who have never interacted in the training data. One is selected as the target person, and an isolated event (with no co-occurrence) is chosen for that person. For the others, we modify the location of their nearby events (i.e., those occurring around the same time) to match the target’s event, creating artificial co-occurrence. This results in a synthetic coordination anomaly, where unrelated agents are made to appear together.
% (3) \textbf{Synthetic Coordination:} We randomly select a group of individuals who have never interacted with one another in the training data and inject a synthetic event by making all of them go to the same location at a similar time. This creates a synthetic coordination anomaly, as the individuals are not supposed to be related, yet are made to appear together.

% (3) Simulated data enables domain experts, such as "red teams," to inject various types of realistic anomalies systematically. This allows for a comprehensive evaluation of the model's ability to detect diverse anomaly types. 

Each dataset is split into train/validation/test by date using a 3/1/4 week ratio. During training, we randomly mask events in each sequence based on a given mask ratio (0.05). For validation and testing, we mask one event at a time, since the model evaluates anomalies on an event-by-event basis.

\begin{table*}[!t]
    \caption{ \method significantly outperforms all baselines at both event-level (left) and agent-level (right) anomaly detection by 3.4\% up to 70\% w.r.t. both AUROC and AUCPR performance. Anomaly detection results depict mean $\pm$ standard dev. over five seeds. Improvement is calculated by $|a - b| / b$, where $a$ is the metric of \method, and $b$ is the metric of the most competitive baseline. Dash (-) denotes the cases for the baselines that cannot provide event-level but only agent-level detection.} 
    \vspace{-0.1in}
    \centering
    \renewcommand\arraystretch{1.1}
    \setlength\tabcolsep{2 pt}
    \resizebox{1 \linewidth}{!}{
    \begin{tabular}{c|cc|cc||cc|cc}
    \toprule
        \multirow{2}{*}{Method} & \multicolumn{4}{c||}{Event-level} & \multicolumn{4}{c}{Agent-level}  \\
        \cline{2-9}
         & \multicolumn{2}{c|}{\trialfourone} & \multicolumn{2}{c||}{\trialfourtwo} & \multicolumn{2}{c|}{\trialfourone} & \multicolumn{2}{c}{\trialfourtwo}  \\
        \cline{2-9}
         & AUROC & AUCPR & AUROC & AUCPR & AUROC & AUCPR & AUROC & AUCPR \\
         \midrule
       % CTLE \cite{2021CTLE}  &  $ \pm $  & $ \pm $  & $ \pm $ &  $ \pm $ & $ \pm $ & $ \pm $ &$ \pm $& $ \pm $\\
        IBAT \cite{IBAT2011}  &  -  & -  & -  &  -  &  0.477$ \pm $ 0.001 & 0.113 $ \pm $ 0.0007 & 0.480 $ \pm $ 0.003 & 0.163 $ \pm $ 0.002\\
        GMVSAE \cite{onlineGMVSAE2020}  & -   &  - &  -  &  -  &  0.492 $ \pm $ 0.004 &  0.119 $ \pm $ 0.003 & 0.496 $ \pm $ 0.004 & 0.167 $ \pm $ 0.001 \\
        ATROM  \cite{ATROM}  &  -  & -  &  -  &  -  &  0.494 $ \pm $ 0.005 & 0.120 $ \pm $ 0.002 & 0.501 $ \pm $ 0.002 & 0.169 $ \pm $ 0.001 \\
        SensitiveHUE \cite{Feng2024HUE} & 0.524 $ \pm $ 0.006 & 0.002 $ \pm $ 0.0002 & 0.529 $ \pm $ 0.005  & 0.004 $ \pm $ 0.0003 & 0.510 $ \pm $ 0.009 & 0.126 $ \pm $ 0.002 & 0.509 $ \pm $ 0.005 & 0.170 $ \pm $ 0.001 \\
        
        % Transformer \cite{Transformer} &  0.643 $ \pm $ 0.000 & 0.009 $ \pm $ 0.000 &  $ \pm $ & $ \pm $ &  0.559 $ \pm $ 0.000 & 0. 155 $ \pm $ 0.000  & $ \pm $ & $ \pm $\\
        
        Transformer-AD \cite{Transformer} &  0.638 $ \pm $ 0.007 & \underline{0.010 $ \pm $ 0.001} &  0.660 $\pm$ 0.007 & \underline{0.016 $\pm$ 0.001} &  \underline{0.558 $ \pm $ 0.006} &  \underline{0.155 $ \pm $ 0.003}  & \underline{0.565 $\pm$ 0.004} & \underline{0.214 $\pm$ 0.003}\\
        
        % Transformer + Friend ID  & 0.634 $ \pm $ 0.000 & 0.011 $ \pm $  0.000 &  $ \pm $ & $ \pm $ &  0.553 $ \pm $ 0.000 & 0.154 $ \pm $ 0.000  & $ \pm $ & $ \pm $\\

        TransformerFriend-AD & \underline{0.643 $ \pm $ 0.007} & 0.009 $ \pm $  0.001 &  0.659 $\pm$ 0.011 & 0.014 $\pm$ 0.001 &  0.554 $ \pm $ 0.007 & 0.151 $ \pm $ 0.005  & 0.562 $\pm$ 0.003 & 0.172 $\pm$ 0.081\\

        % Transformer + Link Score  & 0.605 $ \pm $ 0.000 & 0.003 $ \pm $ 0.000 &  $ \pm $ & $ \pm $ & 0.511 $ \pm $ 0.000  & 0.126 $ \pm $ 0.000  & $ \pm $ & $ \pm $\\

        TransformerLink-AD  & 0.616 $ \pm $ 0.025 & 0.004 $ \pm $ 0.002 &  \underline{0.667 $\pm$ 0.032} & 0.008 $\pm$ 0.005 & 0.526 $ \pm $ 0.011  & 0.126 $ \pm $ 0.006  & 0.531 $\pm$ 0.019 & 0.191 $\pm$ 0.015\\

        % KHAN
        %$P(\hat L)$ $P(\hat M|L)$ one-branch  & 0.742 $ \pm $ 0.000 & 0.014 $ \pm $ 0.000 &  $ \pm $  &  $ \pm $  & 0.563  $ \pm $ 0.000  &  0.163 $ \pm $ 0.000 &  $ \pm $  & $ \pm $ \\ 

        % $P(\hat L)$ $P(\hat M|L)$ one-branch  & \textbf{0.760 $ \pm $ 0.011} & \textbf{0.017 $ \pm $ 0.002} &  $ \pm $  &  $ \pm $  & 0.577  $ \pm $ 0.01  &  0.171 $ \pm $ 0.007 &  $ \pm $  & $ \pm $ \\

         \method & \textbf{0.760 $ \pm $ 0.011} & \textbf{0.017 $ \pm $ 0.002} &  \textbf{0.758 $\pm$ 0.008}  &  \textbf{0.019 $\pm$ 0.002}  & \textbf{0.577  $ \pm $ 0.01}  &  \textbf{0.171 $ \pm $ 0.007} &  \textbf{0.581 $\pm$ 0.003}  & \textbf{0.219 $\pm$ 0.003} \\

        % $P(\hat L)$ $P(\hat M|L)$ one-branch  w/o L  &   $ \pm $  &  $ \pm $  &  $ \pm $   &  $ \pm $  &   $ \pm $  &  $ \pm $  &  $ \pm $ & $ \pm $ \\ 
        % $P(\hat L)$ $P(\hat M|L)$ one-branch  w L  &  $ \pm $  & $ \pm $  &  $ \pm $  &  $ \pm $  &   $ \pm $  &  $ \pm $  &  $ \pm $  & $ \pm $ \\ 

        \hline

         Improvement  & 18.2\% ($\uparrow$)  & 70.0\% ($\uparrow$)  &  13.6\% ($\uparrow$) &  18.7\% ($\uparrow$) & 3.4\% ($\uparrow$)   & 10.3\% ($\uparrow$)   & 3.6\% ($\uparrow$) & 2.3\% ($\uparrow$) \\

    \bottomrule
    \end{tabular}
    }
    \label{tab:anomaly_detection}
    \vspace{-0.1in}
\end{table*}

\subsubsection{Baselines} To evaluate the performance, we implement different types of baselines.

\par \noindent \textbf{Link Prediction Baselines:} We implement the following baselines for link prediction to evaluate the co-occurrence of a candidate pair:

\begin{itemize}[leftmargin=*]
    \item Random: Assigns a random score uniformly sampled from [1, 10] for the co-occurrence of a candidate pair.
    \item HistoryFreq: Uses the historical co-occurrence frequency between two individuals as the score. A higher frequency indicates a stronger likelihood of future co-occurrence.
    % which takes the meeting frequency of the two individuals as the score. Higher frequency means a greater chance of being together.
    \item HistoryDuration: Similar to HistoryFreq, but uses the total co-occurrence duration between two individuals as the score.
\end{itemize}

\par \noindent \textbf{Anomaly Detection Baselines:} There are only a few related methods for human mobility anomaly detection. Thus, it is difficult to find state-of-the-art models for direct comparison. To ensure a comprehensive evaluation, we select baselines from three relevant domains for comparison: 
% \par  (1) Human mobility Learning:
% \begin{itemize}[leftmargin=*]
%     \item CTLE \cite{2021CTLE} learns context and time-aware location embeddings. We then train a transformer to recover the features based on the embeddings and aggregate the prediction errors across features as the anomaly score. Specifically, we compute prediction error for numerical features as $\Delta_f^{\text{num}}= |\widehat{y}_f - y_f|$ and for categorical features as $\Delta_f^{\text{cls}}= 1 - p_{f,c}$ where $c$ is the true class. 
% \end{itemize}

%  We also use Transformer for feature reconstruction and aggregate prediction errors of event features for anomaly scoring. 

\par (1) Trajectory anomaly detection (more details in  Appx.~\ref{appx:settings}): 
\begin{itemize}[leftmargin=*]
    \item IBAT \cite{IBAT2011} detects anomalies using how much the target trajectory can be isolated from other trajectories.
    \item GMVSAE \cite{onlineGMVSAE2020} uses a generative variational sequence autoencoder that learns trajectory patterns with a Gaussian Mixture model and uses the probability of the trajectory being generated as the anomaly score.
    \item  ATROM \cite{ATROM} uses variational Bayesian methods and correlates trajectories with possible anomalous patterns with the probabilistic metric rule.
\end{itemize}

\par (2) Multi-variate time series anomaly detection: 
\begin{itemize}[leftmargin=*]
    \item SensitiveHUE \cite{Feng2024HUE} uses a probabilistic network by implementing both reconstruction and heteroscedastic uncertainty estimation, and uses both terms as anomaly score.
\end{itemize}

\par (3) Human mobility learning: Transformer-based~\cite{Transformer} models have become the dominant approaches for human mobility learning \cite{2021CTLE,chen2021robust,jiang2023self,yang2023lightpath}. To adapt them for anomaly detection, we implement the following three variants:
%Transformer~\cite{Transformer} has been the de facto most popular architecture for human mobility learning \cite{2021CTLE,chen2021robust,jiang2023self,yang2023lightpath}. We implement three variants of it for the anomaly detection task:
% we adopt the three variants of it to compare with the proposed method.
\begin{itemize}[leftmargin=*]
    \item Transformer-AD \cite{Transformer} uses a vanilla Transformer encoder applied to the same input features in~\eqnref{eq:input-features}. It is pre-trained using the reconstruction loss in~\eqnref{eq:node-loss}, and the anomaly score is computed from the reconstruction error as in~\eqnref{eq:ad-node-reconstruction}.

    %, which takes the vanilla Transformer as encoder, which takes the same feature in Eq.~\eqref{eq:input-features}, and is pre-trained with reconstruction loss as in Eq.~\eqref{eq:node-loss}. The anomaly score is based on the reconstruction error as in Eq.~\eqref{eq:marker-reconstruction}.

    \item TransformerFriend-AD uses the embeddings for neighboring individuals from ($\mathcal{R}_w^u$ defined in Sec.~\ref{sec:problem-definition-ces}) as additional features for each event’s input features defined in ~\eqnref{eq:input-features}. These user embeddings are learned via an embedding table. The architecture and anomaly scoring follow the same setup as Transformer-AD.
    
    % which adds additional neighbor ($\mathcal{R}_w^u$ defined in Sec.~\ref{sec:problem-definition-ces}) embeddings (learned by a user embedding table) as features for each event features in  Eq.~\eqref{eq:input-features}. The architecture and anomaly score are the same as Transformer-AD.

    \item TransformerLink-AD uses the same architecture as Transformer-AD but includes additional link anomaly scores (i.e., $f_{\text{AS}}(e)_{\rm obs-link}$ and $f_{\text{AS}}(e)_{\rm not-obs-link}$) alongside the node reconstruction score as in ~\eqnref{eq:ad-score-node-link}. To calculate the link scores, the item $S_{\rm link}(\cdot)$ in ~\eqnref{eq:ad_score_obs_link} and~\eqnref{eq:ad_score_not_obs_link} is defined as the likelihood of co-occurrence between two individuals, and computed from their meeting frequency in the training data.
    % , defined as the likelihood of co-occurrence between two individuals,
    % $S_{\rm link}(\cdot)$, 
    %The chance that two people are coordinated $S_{\rm link}(\cdot)$ is computed by the frequency that they meet in the training data.
    %it is equipped with anomaly score as in Eq.~\eqref{eq:loss_total},
    %which adds additional link anomaly scores instead of the node reconstruction anomaly score compared with Transformer-AD. While the chance that two people are together $S_{\rm link}(\cdot)$ is computed by the frequency that they meet in the training data.
 
\end{itemize}

Note that IBAT, GMVSAE, and ATROM cannot flag event-level anomalies. Therefore, we only report their performance in agent-level anomaly detection. We did not include graph-based solutions since they do not model sequential trajectories or spatial information explicitly, making them unsuitable for direct comparison. 

%with our method.
% not using space, not using future information, not suitable for comparison

% \input{tables/tab_ad}

%\vspace{-.1in}
\subsubsection{Model Configurations} 
We refer to \appref{appx:settings} for detailed training settings and model hyperparameters.

%\vspace{-.1in}
\subsubsection{Evaluation Metrics}
We report AUCROC and AUCPR, respectively the area under the ROC and Precision-Recall curves, to evaluate anomaly detection performance.
We also evaluate link prediction using three metrics: (1) HR$@k$ (Hit Rate): Measures whether the true positive link appears among the top-$k$ predicted candidates;
(2) MRR (Mean Reciprocal Rank): Measures how early the first positive link appears in the ranked list; (3) Jaccard Similarity@$\alpha$: Computes the similarity between the predicted  and ground truth link sets, using a threshold $\alpha$ to select top-scoring predictions. We report $JS@\alpha_{\rm best}$, where $\alpha_{\rm best}$ is selected by uniformly searching over 50 values in [0,1], and choosing the one that maximizes the F1 score on the validation set. We refer to \appref{appx:metric} for  details.

% \begin{table*}[htbp]
% \centering
% \small
% \caption{\centering Link prediction results. Higher HR@$k$, MRR, and JS@$\alpha_{\rm best}$ means better performance.}
% \vspace{-0.1in}
% \renewcommand\arraystretch{1.1}
% \setlength\tabcolsep{2 pt}
% \resizebox{0.9 \textwidth}{!}{
% \begin{tabular}{c|ccccc||ccccc}
% \toprule
% \multirow{2}{*}{Method} 
% & \multicolumn{5}{|c||}{\trialfourtwo} & \multicolumn{5}{|c}{\trialfourone} \\
% \cline{2-11}
% & HR@1($\uparrow$) & HR@2($\uparrow$) & HR@3($\uparrow$) & MRR($\uparrow$) & JS@$\alpha_{\rm best}$ & HR@1($\uparrow$) & HR@2($\uparrow$) & HR@3($\uparrow$) & MRR($\uparrow$) & JS@$\alpha_{\rm best}$ \\
% \midrule

% Random & & & & & & 0.736 & 0.947 & 0.977 & 0.672 & 0.27 \\
% HistoryFreq & & & & & & \underline{0.887} & \underline{0.977} & \underline{0.990} & \underline{0.747} & \underline{0.74} \\
% HistoryDuration & & & & & & 0.866 & 0.973 & 0.987 & 0.738 & 0.74 \\

% \method & & & & & & \textbf{0.931} & \textbf{0.986} & \textbf{0.993} & \textbf{0.771} & \textbf{0.78} \\

% \bottomrule
% \end{tabular}
% }

% \label{tab:link_prediction}
% \end{table*}

\begin{table*}[tbp]
\centering
\small
\caption{\centering \method significantly improves link prediction over simple alternatives w.r.t. all metrics; HR@$k$, MRR, and JS@$\alpha_{\rm best}$.}
\vspace{-0.1in}
\renewcommand\arraystretch{1.1}
\setlength\tabcolsep{2pt}
\resizebox{0.8\textwidth}{!}{
\begin{tabular}{c|ccccc||ccccc}
\toprule
\multirow{2}{*}{Method}
  & \multicolumn{5}{|c||}{\trialfourone}
  & \multicolumn{5}{|c}{\trialfourtwo} \\
\cline{2-11}
  & HR@1($\uparrow$) & HR@2($\uparrow$) & HR@3($\uparrow$) & MRR($\uparrow$) & JS@$\alpha_{\rm best}$
  & HR@1($\uparrow$) & HR@2($\uparrow$) & HR@3($\uparrow$) & MRR($\uparrow$) & JS@$\alpha_{\rm best}$ \\
\midrule
Random
  & 0.736            & 0.947            & 0.977            & 0.672           & 0.270
  & 0.734            & 0.947            & 0.976            & 0.643           & 0.566    \\
HistoryFreq
  & \underline{0.887} & \underline{0.977} & \underline{0.990} & \underline{0.747} & \underline{0.740}
  &  \underline{0.883}            &  \underline{0.976}            &  0.989            &  \underline{0.749}            & \underline{0.617}     \\
HistoryDuration
  & 0.866            & 0.973            & 0.987            & 0.738           & 0.740
  & 0.865            & 0.971            & \underline{0.989}            & 0.741           & 0.603     \\
\method
  & \textbf{0.931}   & \textbf{0.986}   & \textbf{0.993}   & \textbf{0.771}  & \textbf{0.780}
  &  \textbf{0.945}           &  \textbf{0.990}           &  \textbf{0.996}           &  \textbf{0.783}          & \textbf{0.820}     \\
\bottomrule
\end{tabular}
}
\label{tab:link_prediction}
\vspace{-0.1in}
\end{table*}

% $P(\hat L)$ $P(\hat M|L)$ one-branch (KNAN) & & & & & & \textbf{0.931} & \textbf{0.986} & \textbf{0.993} & \textbf{0.771} & \textbf{0.78} \\

%$P(\hat M, \hat L)$  &  -  &  -  &  -  &  -  & -  &  -  &  -  &  -  \\ \midrule

% $P(\hat L)$ $P(\hat M|L)$ one-branch &  0.53  &  0.76  &  0.87  &  0.71  &  0.26  &  0.50  &  0.69  &  0.50  \\ \midrule

%$P(\hat L)$ $P(\hat M|L)$ two-branch &  0.67  &  0.91  &  0.97  &  0.82  &  0.51  &  0.79  &  0.91  &  0.71  \\

\subsection{Q1: Anomaly Detection Performance}

\tabref{tab:anomaly_detection} reports the performance of all methods on both event-level and agent-level anomaly detection. \method significantly outperforms all baselines, improving AUCROC by 13.6\%-18.2\% and AUCPR by 18.7\%-70\% for event-level detection compared to the most competitive baseline. 
Trajectory-based anomaly detection methods such as IBAT, GMVSAE, and ATROM are restricted to agent-level detection since they compute anomaly scores based on the likelihood of trajectory (i.e., GPS coordinate sequences). These methods perform poorly in our setting because they fail to model the rich spatio-temporal features and semantic contexts in human mobility.
%Trajectory-based anomaly detection methods such as IBAT, GMVSAE, and ATROM evaluate the trajectory (i.e., GPS coordinate sequences) probability as the anomaly score, which are restricted to agent-level detection, and exhibit poor performance because they overlook the complex spatio-temporal features and semantics embedded in human mobility. 
SensitiveHUE improves upon trajectory-based methods by modeling numerical event features as a multi-variate time series and incorporating learned uncertainty into the anomaly score. However, it  lacks the ability to model joint spatial-temporal interactions, which are essential for capturing meaningful mobility patterns. Its inability to handle categorical features further limits its effectiveness.  

TransformerFriend-AD adds neighbor embeddings to enrich each event representation with information about who the individual is interacting with. This does not consistently improve performance. In some cases, it even causes a slight drop (e.g., AUCPR decreases from 0.01 to 0.009 in event-level detection on \trialfourone). This suggests that naively incorporating neighbor information is insufficient and that collective behavior modeling requires more architectural development. Moreover, TransformerLink-AD, which introduces a link-based anomaly score derived from historical meeting frequency, also underperforms compared to Transformer-AD. This indicates that frequency alone cannot capture the complexity of spatio-temporal human interactions. In contrast, \method achieves the best performance due to two key design choices: (1) the two-stage attention mechanism that effectively captures both individual and collective behavior patterns; and (2) the designed anomaly scoring function, which detects diverse anomaly types by using both feature and link reconstruction signals.

% By doing so, we find that the performance does not constantly improve but with a slight performance drop in some cases (e.g., from 0.01 to 0.009 for AUCPR of event-level anomaly detection in \trialfourone). Such observation shows that it is non-trivial to model the collective behavior, since directly adding the neighbor's information is not an effective way to boost the performance. 
% Moreover, TransformerLink-AD (introduces the link anomaly score based on meeting frequency) performs worse than Transformer-AD, which shows that frequency-based method cannot well capture the complex spatio-temporal patterns in people's interactions. In contrast, \method achieves the best performance mainly benefited by (1) the two-stage attention that captures both individual and collective behavior patterns; (2) the designed anomaly function is effective in capturing different types of anomalies.

%As for the Transformer-based methods, 

% compared to \method. 

% Although SensitiveHUE improves upon earlier methods by incorporating uncertainty into multivariate time series modeling, it focuses solely on temporal dependencies and fails to capture the joint spatial-temporal interactions critical for understanding mobility patterns. Additionally, its inability to handle categorical features further constrains its effectiveness compared to \method. 

% \subsection{Q3: Ablation Study}

% \begin{itemize}
%     \item remove collective correlation modeling
%     \item remove temporal dependencies modeling 
%     \item about anomaly scoring
    
% \end{itemize}

\vspace{-0.1in}
\subsection{Q2: Link Prediction Performance}
% current baseline + potential 
% basic baseline: lstm, Transformer, CTLE, LightPath
% our method

% We report the performance of different methods in the validation dataset since it does not contain any other anomalous data. 

We evaluate the link prediction performance using the validation dataset, which contains no anomalies. \tabref{tab:link_prediction} reports the performance of various link prediction methods. \method consistently outperforms all baselines across all metrics, demonstrating its superior ability to identify plausible links. In \trialfourone, our model achieves top performance with HR@1 of 0.931, MRR of 0.771, and JS@$\alpha_{\rm best}$ of 0.78, improving upon History-meet-freq by 4.4\%, 3.2\%, and 5.4\%, respectively. These results confirm that \method captures the underlying link structure more effectively and provides more accurate and discriminative predictions than baselines relying on simple heuristics such as historical meeting frequency or duration.

\vspace{-0.1in}
\subsection{Q3: Ablation Study}

\begin{figure}[htbp]
    \centering
    \includegraphics[width=0.9 \linewidth]{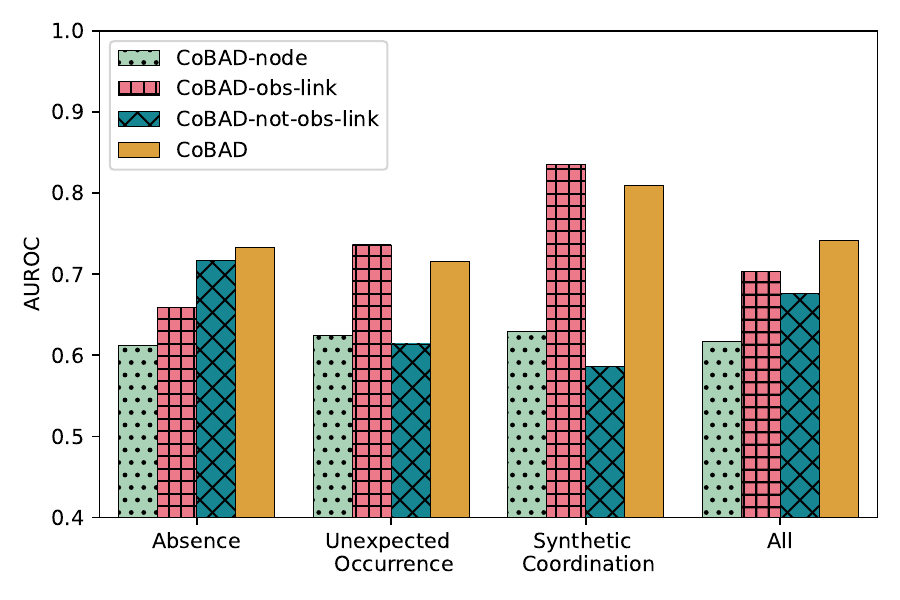}
    \vspace{-0.15in}
    \caption{ \method can detect collective anomalies of various types. Ablation study of \method variants on different anomalies: adding \method-obs-link loss to node reconstruction loss  boosts performance significantly on unexpected occurrence and synthetic coordination anomalies, while  \method-not-obs-link loss is most effective for absence anomalies.  }
   % \vspace{-0.1in}
    \label{fig:ablation-AUCROC}
\end{figure}

To better understand the contribution of different anomaly score components, we compare different variants of \method under three types of injected anomalies. The four variants are:
($i$) \method-node: Uses only the node reconstruction score defined in~\eqnref{eq:ad-node-reconstruction}; ($ii$) \method-obs-link: Extends \method-node by adding the observed-link anomaly score from~\eqnref{eq:ad_score_obs_link}; 
 ($iii$) \method-not-obs-link: Extends \method-node by adding the unobserved-link anomaly score from \eqnref{eq:ad_score_not_obs_link};
and ($iv$) \method: the full model, incorporating all three anomaly score components. 

%The types of injected anomalies are: (1) Absence: One agent is removed from a co-occurrence event group by changing its POI type and location to simulate a missing participant;
%(2) Unexpected occurrence: An unrelated agent is injected into an existing co-occurrence group, disrupting its original semantics; (3) Synthetic Coordination: A synthetic co-occurrence event is created among individuals who have never interacted, introducing a new and artificial collective pattern.
%These perturbations are designed to challenge the model’s understanding of different collective patterns.
%coordination structures. 

% (see also Appx.~\tabref{tab:ad_insert_event} for more results)
In ~\figref{fig:ablation-AUCROC} \method-node, which uses only the node loss, consistently performs the worst across most anomaly types, highlighting its limitations in capturing collective anomalies.
Adding the observed-link loss (i.e., \method-obs-link) significantly improves performance, especially for Unexpected Occurrence and Synthetic Coordination anomalies, where it achieves the highest AUCROC (0.736 and 0.835, respectively). This aligns with its design goal: penalizing unexpected observed links. Conversely, adding the not-observed-link loss (i.e., \method-not-obs-link) is most effective for Absence anomalies, yielding the highest AUCROC (0.717). This demonstrates its ability to detect missing but expected interactions. Finally, our full model—combining both link-based losses with the node loss—achieves the best overall performance on the mixed anomaly set, % (AUCROC of 0.742 and AUCPR of 0.014), 
underscoring the effectiveness of joint anomaly scoring to capture diverse collective anomalies.
% Predictable:
% 1. Absence: For a coordinates event group, One agent does not show up to the coordinates event (changing its POI type or removing that event completely)
% 2. Stranger: An unrelated agent joins the coordinated event
% 3. Group anomaly: all agents in that coordinated group change POI type or time

% OOD:
% 1. Synthetic Coordination: For agents without any relation or coordinated events, select a few and create one for them

\begin{figure*}[!t]
    \centering
    \includegraphics[width=\linewidth]{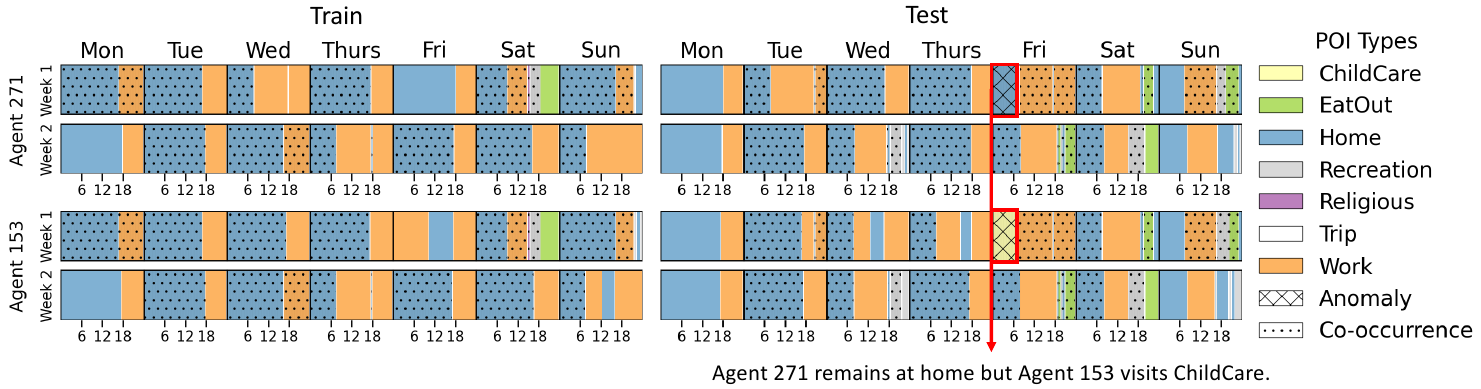}
     \vspace{-0.2in}
    \caption{(best viewed in color) Case study: A detected “absence anomaly” (i.e., high score in the missing link). Visualization shows agent behavior during the training period (left) and the test period (right). During the train period, agents 271 (target agent) and 153 are family members who typically co-occur at home during late-night to early-morning hours. Co-occurrence ($\cdot$) marks events where all agents are present together. Anomaly ($\times$) marks the target event. During the test period, however, Agent 271 remains at home, but Agent 153 visits a ChildCare location. This deviation creates an absence anomaly for agent 271, which is effectively detected by a high score in the missing link between 153 and 271.}
    \label{fig:absence_case}
     \vspace{-0.1in}
\end{figure*}

\begin{figure*}[!t]
    \centering
    \includegraphics[width=\linewidth]{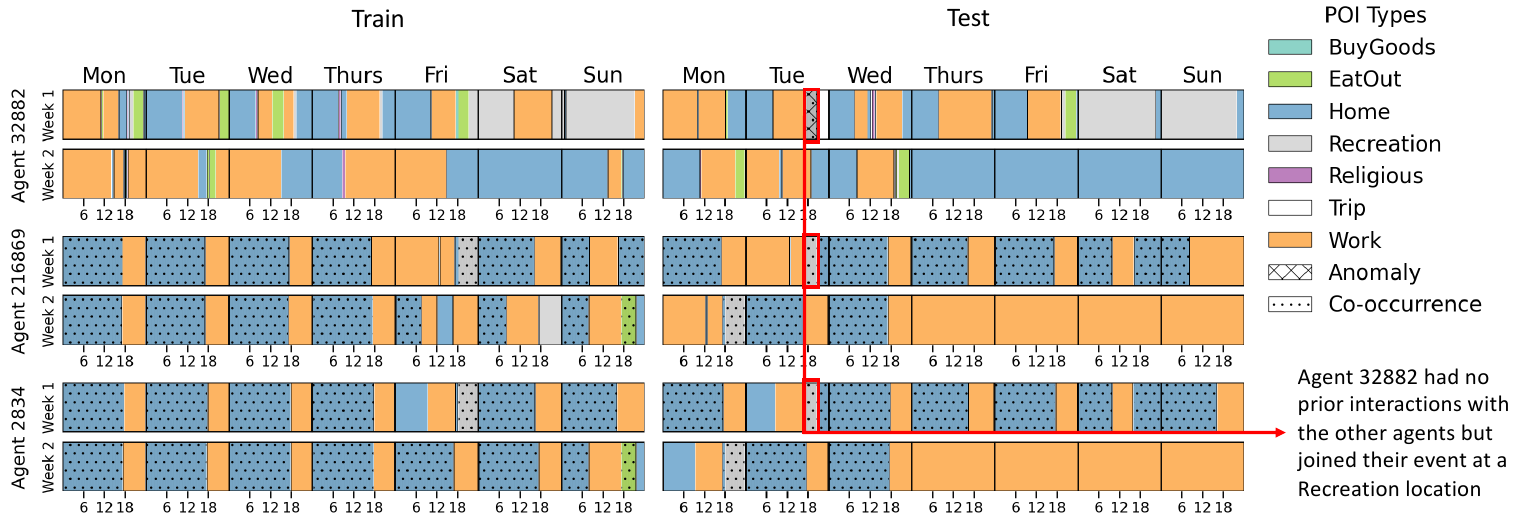}
     \vspace{-0.2in}
    \caption{(best viewed in color) Case study: A detected "unexpected occurrence anomaly" (i.e., high score in the observed link). The figure visualizes agent behavior during the training period (left) and testing period (right). During training, agents 216869 and 2834, who are family members, consistently visit the same locations together. In contrast, agent 32882 has no prior interactions with them. Co-occurrence ($\cdot$) denotes events where agents 216869 and 2834 appear together. Anomaly ($\times$) marks the target anomaly event. During testing, agent 32882 unexpectedly joins the other two agents at a location he had previously visited alone. This deviation is flagged as an anomaly by the observed link module due to the unexpected co-occurrence.
    }
    \label{fig:stranger_case}
\end{figure*}

\vspace{-0.1in}
\subsection{Q4: Case Studies}
To gain an intuitive understanding of the anomalies detected by our model and to assess whether the link prediction module captures collective anomalies, we analyze two representative cases: one absence anomaly and one unexpected occurrence anomaly. ~\figref{fig:absence_case} illustrates an absence anomaly. We visualize the event sequences of the target agent and his related agents during both the training and testing periods to highlight the expected behavioral patterns and the input context surrounding the anomaly. For clarity, we present a simplified scenario involving only two related agents. In this case, the target agent (271) follows his typical routine and does not exhibit any anomalous behavior individually. However, the absence of their usual co-occurrence with a related agent (who visits a different location alone while 271 remains at the usual home location) results in a high missing link score. This disruption in the expected co-occurrence pattern signals an anomaly, and the model successfully detects it. ~\figref{fig:stranger_case} shows an example of the unexpected occurrence anomaly. Agent 32882 appears at a location alongside agents with whom he has no prior interactions. The observed links between them are unexpected and hence signal the anomaly.
In summary, \method~ effectively utilizes link prediction to detect collective anomalies. Moreover, the resulting scores enhance interpretability by indicating the potential type of anomaly.

% \begin{figure}[htbp]
%     \centering
%     \includegraphics[width=1 \linewidth]{img/time.pdf}
%     \caption{Left: training time per sample w.r.t  $L$ when $N$=5, right: training time per sample w.r.t  $N$ when $L$=10. Common setting: batch size = 128, $D$=32.}
%     \label{fig:time_N_L}
% \end{figure}

% \begin{figure}[htbp]
%     \centering
%     \includegraphics[width=1 \linewidth]{img/memory.pdf}
%     % \caption{GPU Memory Usage per batch w.r.t $N$ and $L$. Setting: batch size = 128, $D$=32.}
%     \caption{Left: GPU Memory Usage per batch w.r.t  $L$ when $N$=5, right: GPU Memory Usage per batch w.r.t  $N$ when $L$=10. Common setting: batch size = 128, $D$=32.}
%     \label{fig:mem_N_L}
% \end{figure}

\vspace{-0.1in}
\subsection{Q5: Scalability}
For brevity, we show the training time and GPU memory usage of \method in \appref{sec:scale}, empirically demonstrating that it scales near-linearly with increasing input size, 
making it efficient and practical for large-scale learning tasks.

\section{Conclusion}
We proposed \method, a new collective behavior model for human mobility that explicitly captures spatial, temporal, as well as relational patterns toward anomaly detection. \method employs a novel two-stage attention mechanism to jointly model spatiotemporal dependencies within individual sequences as well as relational dependencies between individuals. As such, \method detects anomalies based on both event attributes and their co-occurrence relations. 
Extensive experiments on industry-scale datasets showed that \method outperforms existing baselines in both event-level and agent-level collective anomaly detection. We expect our work and open-source codebase to foster further research on modeling human mobility through the lens of social interactions and collective dynamics.

\clearpage

\section{Acknowledgement}
This work is supported by the Intelligence Advanced Research Projects Activity (IARPA) via Department of Interior/Interior Business Center (DOI/IBC) contract number 140D0423C0033. The U.S. Government is authorized to reproduce and distribute reprints for Governmental purposes notwithstanding any copyright annotation thereon. Disclaimer: The views and conclusions contained herein are those of the authors and should not be interpreted as necessarily representing the official policies or endorsements, either expressed or implied, of IARPA, DOI/IBC, or the U.S. Government.

\bibliographystyle{ACM-Reference-Format}
\bibliography{ref}

\appendix

\section{Dataset Details}
\label{sec:data}
The summary statistics of the datasets are in~\tabref{tab:stat_test_data}.

\begin{table}[htbp]%
	\centering
	\caption{Dataset statistics.}
 \vspace{-0.1in}
	\setlength\tabcolsep{2pt}  % Increased padding for better readability
	\resizebox{0.85 \linewidth}{!}{  % Ensure it fits within the page width
		\begin{tabular}{lcc}
			\toprule
			& \trialfourone &\trialfourtwo \\
			\midrule
			\#Agents & 20,000 & 20,000 \\
			\#Anomalous Agents & 2,415 & 3,379 \\
                Ratio of Anomalous Agents & 12.1\% & 16.9\%\\ \midrule
			
			\#Events (Test Period) & 1,770,428 & 1,762,687\\ % test
			\#Anomalous Events & 2,595 & 3,635\\ 
                Ratio of Anomalous Events & 0.15\% & 0.21\%\\ \midrule
                %\#Events & 1,549,104 & 1,794,537 & 1,706,468 \\ % only test
                % Ratio of Anomalous Events & 0.0006 & - & 0.002 \\ \hline
                Avg $N$ & 2.45 & 2.37 \\
                Avg $L$ & 7.89 & 7.88\\
                x range (km) & [-4.6, 4.6] & [-4.6, 4.6]\\
                y range (km) & [-5.6, 5.6] &  [-5.6, 5.6]\\
                Avg of stay duration (min) & 543 & 547 \\
                Avg of start time (min) & 764 & 765 \\ 

                \#{\rm poi} &  14 & 14 \\

                \midrule

                Time Span & 66 days & 66 days \\ 
                City & Tokyo  & Tokyo \\
                % Public/Private & Private  & Private \\     
			\bottomrule
		\end{tabular}
	}
	\label{tab:stat_test_data}
\end{table}

\section{Model Configurations} \label{appx:settings}

 For all deep models, we search the hyperparameters in a given model space, and each model's best performance is reported in the experiments. The batch size of each epoch is set to 128. The learning rate of the Adam optimizer starts from 1e-3 with a weight decay of 1e-05. 
 % For our proposed \method, we use dropout layers after each feature's embedding layer with a dropout rate of 0.05, which randomly drops a subset of neurons during both training and inference. 
 The hyperparameter space of \method is as follows:  mask ratio for pre-training is searched from [0.05, 0.3]; while embedding size $D$ for the transformer is searched from [32, 64, 128]. The number of negative edges for a positive edge is set to 5, the number of TSA layers $M$ is set to 1, and the number of heads $H$ in the attention mechanism is set to 4.
 Each event sequence is set to span three consecutive
days, and $\delta$ is set to 40 meters to depict co-occurrence between events.

% \blue{Haomin: edit this part --> The layer $M_1$ for the feature-level transformer is set to 1 for simplicity, and the layers $M_2$ for the event-level transformer are searched from [3, 4, 5]. }

For the baseline methods in trajectory and time series anomaly detection, we describe the computation of anomaly scores and the hyperparameter space in our experiments as follows. 
\begin{itemize}[leftmargin=*]

    \item IBAT \cite{IBAT2011} determines anomaly scores based on how easily a trajectory can be isolated from others, where easier isolation indicates a rarer pattern and higher anomalousness.  The number of running trials is searched from [50, 100, 200, 300], and the subsample size is searched from [64, 128, 256, 512].

    \item GMVSAE \cite{onlineGMVSAE2020} computes anomaly score as the probability of the trajectory being generated from the learned patterns. The embedding size is searched from [256, 512], and the number of clusters is searched from [5, 10]. 

    \item ATROM \cite{ATROM} assigns anomaly scores based on the probability of a trajectory being classified into predefined anomaly categories. The embedding size is searched from [64, 128, 512].

    \item SensitiveHUE \cite{Feng2024HUE} computes anomaly score for time $t$ and channel $s$ as $S(t,s) = \frac{(\widehat{\mu}_{ts} - X_{ts})^2}{2 \widehat{\sigma}_{ts}^2} + \frac{1}{2} \ln \widehat{\sigma}_{ts}^2$, where $\widehat{\mu}_{ts}$ is the reconstruction, $X_{ts}$ is the input, and $\widehat{\sigma}_{ts}^2$ is the estimated uncertainty. The final score across multiple channels is $\max_{s} \tilde{S}(t,s) = \frac{S(t,s) - \text{Median}(s)}{\text{IQR}(s)}$. The embedding size is searched from [64, 128], and the number of layers is [1, 2].

    \item For Transformer-based baselines, including Transformer-AD, TransformerFriend-AD and TransformerLink-AD, the hidden dimension is searched from [32, 64, 128]. For TransformerFriend-AD, the embedding size of user table is set to 10.
\end{itemize}
  
For all baselines, the agent-level anomaly score is defined as the maximum event score across all events for a given agent.

\section{Evaluation Metrics} \label{appx:metric}

For each target event, we evaluate link prediction by ranking all candidate individuals. Let $P = \{p_1, \dots, p_N\}$ be the input candidate set. The corresponding ground truth labels are given by $Y = \{y_1, \dots, y_N \}$ where $y_i \in \{0, 1\}$ and $y_i=1$ denotes that person $p_i$ co-occurs with the target individual. The model outputs a score $s_i$ representing the likelihood that
$p_i$ co-occurs with the target person, yielding a score list $S = \{s_1, \dots, s_N\}$. Sorting these scores gives a ranked list of candidates $\bm{\widehat{\pi}} = \{ {\widehat{\pi}}_1, \dots, {\widehat{\pi}}_N \}$, where $\widehat{\pi}_i = j$ means the $i$-th ranked person is $p_j$. Based on the above setup, the metrics are computed as below:

\vspace{.1in}
\par \noindent  \textbf{HR@$k$} measures the fraction of true positive individuals among the top-$k$ predictions:
\begin{equation} 
    \textbf{\rm HR@}k= 
    \frac{|\{i | i\in [1,k] ~\&~ y_{\widehat \pi_i}=1\}|}{k}.
    \label{eq_hit_rate}
\end{equation}

\vspace{.1in}
\par \noindent  \textbf{MRR} (Mean Reciprocal Rank) measures how early the positive link appears in the ranked list. Let $P_{\rm gt} = \{p_i | y_i =1 \}$ be the set of ground truth positive individuals. MRR is formulated as
\begin{equation}
    M R R=\frac{1}{|P_{\rm gt} |} \sum_{p_i \in P_{\rm gt}} \frac{1}{{\rm rank}(p_i)}\;,
\end{equation}
where ${\rm rank}(p_i)$ is the position of $p_i$ in the ranked list.

\vspace{.1in}
\par \noindent \textbf{Jaccard Similarity (JS)} measures the similarity between the ground truth and prediction. Given a threshold $\alpha$, we define the set of predicted individuals as $P_{\rm hat} = \{p_i | s_i \geq \alpha \}$. JS@$\alpha$ is defined as
\begin{equation}
    {JS@\alpha}=\frac{|P_{\rm gt} \cap P_{\rm hat}|}{|P_{\rm gt} \cup P_{\rm hat}|}\;.
\end{equation}

When both $P_{\rm gt}=\phi$ and $P_{\rm hat}=\phi$, $JS@\alpha = 1$. In 
In the experiment, we report $JS@\alpha_{\rm best}$, where $\alpha_{\rm best}$ is selected by uniformly searching over 50 values within the valid range of $\alpha$, and choosing the one that yields the highest F1 score on the validation set.

\section{Scalability}
\label{sec:scale}

\begin{figure}[H]
    \centering
    \includegraphics[width=1 \linewidth]{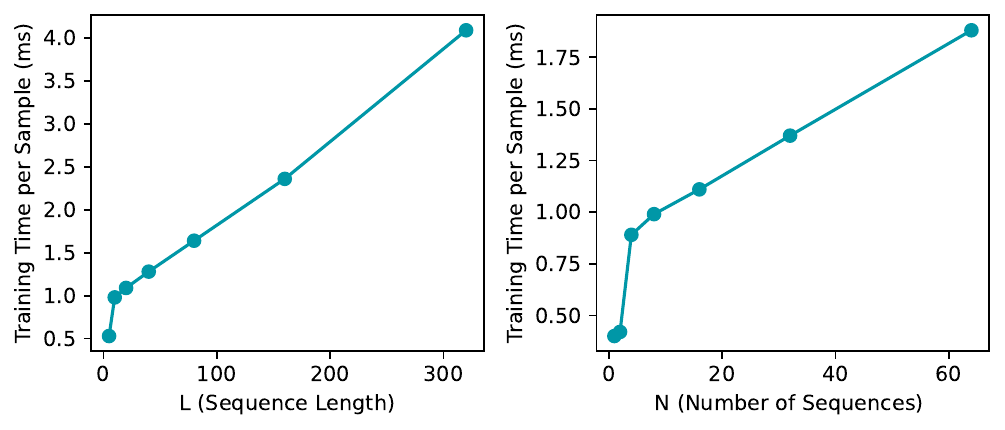}
     \vspace{-.2in}
    \caption{\method scales near-linearly with input size. Left: training time per sample w.r.t  $L$ when $N$=5, right: training time per sample w.r.t  $N$ when $L$=10. Common setting: batch size = 128, $D$=32.}
    \label{fig:time_N_L}
\end{figure}

\begin{figure}[b]
    \centering
    \includegraphics[width=1 \linewidth]{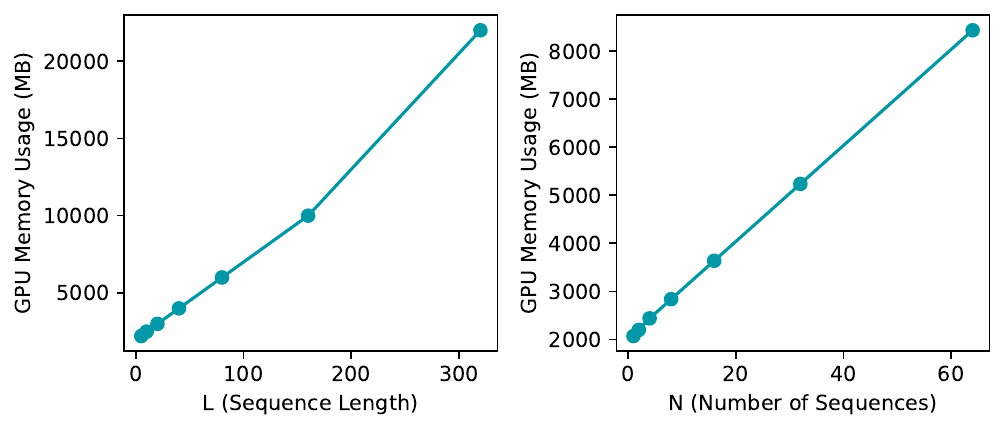}
    \vspace{-.2in}
    \caption{\method uses memory efficiently, proportionate to input size.  Left: GPU Memory Usage per batch w.r.t  $L$ when $N$=5, right: GPU Memory Usage per batch w.r.t  $N$ when $L$=10. Common setting: batch size = 128, $D$=32.}
    \label{fig:mem_N_L}
\end{figure}

To evaluate the scalability of the model, we report the training time per sample and the GPU memory usage under varying numbers of neighbors $N$ and sequence lengths $L$, with a fixed batch size (128) and embedding dimension D = 32.
The results show that both the training time per sample (see ~\figref{fig:time_N_L}) and GPU memory usage (see ~\figref{fig:mem_N_L}) scale approximately linearly with $L$ and $N$. This demonstrates that the proposed model maintains favorable scalability in terms of both computational cost and memory consumption, making it efficient and practical for large-scale learning tasks.

All models are trained on 1 GPU of NVIDIA RTX A6000.

\end{document}